\documentclass{article}
\usepackage{PRIMEarxiv}
\usepackage{microtype}

\usepackage[utf8]{inputenc}
\usepackage[T1]{fontenc}
\usepackage{amsmath,amssymb,amsfonts,amsthm}
\usepackage{graphicx}
\usepackage{booktabs}
\usepackage{tabularx}
\usepackage{multirow}
\usepackage{xcolor}
\usepackage[numbers]{natbib}
\usepackage{hyperref}
\usepackage{cleveref}
\usepackage{algorithm}
\usepackage{algorithmic}
\usepackage{enumitem}
\usepackage{float}
\usepackage{tikz}
\usetikzlibrary{shapes.geometric, arrows.meta, positioning, backgrounds, fit, calc}
\usepackage{pgfplots}
\pgfplotsset{compat=1.18}
\usepackage[most]{tcolorbox}

\definecolor{cAction}{HTML}{CCE5FF}     % light blue
\definecolor{cEntity}{HTML}{FFD9B3}     % light orange
\definecolor{cLocation}{HTML}{C8E6C9}   % light green
\definecolor{cTime}{HTML}{E1BEE7}       % light purple
\definecolor{cQuantity}{HTML}{FFCDD2}   % light red
\definecolor{cCompare}{HTML}{B2EBF2}    % light cyan
\definecolor{cMechanism}{HTML}{FFF9C4}  % light yellow

\definecolor{cActionD}{HTML}{1565C0}
\definecolor{cEntityD}{HTML}{E65100}
\definecolor{cLocationD}{HTML}{2E7D32}
\definecolor{cTimeD}{HTML}{6A1B9A}
\definecolor{cQuantityD}{HTML}{C62828}
\definecolor{cCompareD}{HTML}{00838F}
\definecolor{cMechanismD}{HTML}{F9A825}

\definecolor{promptBlue}{HTML}{E3F2FD}
\definecolor{promptBlueBorder}{HTML}{1565C0}
\definecolor{posGreen}{HTML}{E8F5E9}
\definecolor{posGreenBorder}{HTML}{2E7D32}
\definecolor{negRed}{HTML}{FBE9E7}
\definecolor{negRedBorder}{HTML}{C62828}

\newtcolorbox{promptbox}[1][]{enhanced, breakable,
  colback=promptBlue, colframe=promptBlueBorder, coltitle=white,
  fonttitle=\bfseries\small, left=6pt, right=6pt, top=4pt, bottom=4pt,
  boxrule=0.8pt, arc=2pt, title={#1}, before upper={\small}}

\newtcolorbox{positivebox}[1][]{enhanced, breakable,
  colback=posGreen, colframe=posGreenBorder, coltitle=white,
  fonttitle=\bfseries\small, left=6pt, right=6pt, top=4pt, bottom=4pt,
  boxrule=0.8pt, arc=2pt, title={#1}, before upper={\small}}

\newtcolorbox{negativebox}[1][]{enhanced, breakable,
  colback=negRed, colframe=negRedBorder, coltitle=white,
  fonttitle=\bfseries\small, left=6pt, right=6pt, top=4pt, bottom=4pt,
  boxrule=0.8pt, arc=2pt, title={#1}, before upper={\small}}

\newcommand{\hlc}[2]{{\fboxsep=0pt\colorbox{#1}{\kern1pt#2\kern1pt}}}

\newcommand{\goldenfact}{\textsc{Golden Fact}}
\newcommand{\goldenfacts}{\textsc{Golden Facts}}
\newcommand{\farmerchat}{\textsc{Farmer.Chat}}
\newcommand{\dgeval}{\textsc{DG-Eval}}
\providecommand{\url}[1]{\texttt{#1}}
\newcommand{\doi}[1]{doi: #1}
\title{Fine-Tuning and Evaluating Conversational AI for Agricultural Advisory}

\author{
   \\
  \textbf{Digital Green}\\
Sanyam Singh,
Naga Ganesh,
Vineet Singh,
Lakshmi Pedapudi,
Ritesh Kumar,
SSP Jyothi, \\
Archana Karanam,
Waseem Pasha,
Ekta Kumari,
C. Yashoda,
Mettu Vijaya Rekha Reddy,
Shesha Phani Debbesa,
Chandan Dash \\[6pt]
\texttt{\{sanyam, naga, vineet, lakshmi, ritesh, jyothi,} \\
\texttt{archana, waseempasha, ekta, yashoda, rekha, shesha, chandan\}@digitalgreen.org}
}

\begin{document}

\maketitle

% Abstract
% Abstract
\begin{abstract}
Large Language Models show promise for agricultural advisory, yet vanilla models exhibit unsupported recommendations, generic advice lacking specific, actionable detail, and communication styles misaligned with smallholder farmer needs. In high-stakes agricultural contexts, where recommendation accuracy has direct consequences for farmer outcomes, these limitations pose challenges for responsible deployment.

We present a hybrid LLM architecture that decouples factual retrieval from conversational delivery: supervised fine-tuning with LoRA on expert-curated \goldenfacts{} (atomic, verified units of agricultural knowledge) optimizes fact recall, while a separate stitching layer transforms retrieved facts into culturally appropriate, safety-aware responses. Our evaluation framework, \dgeval{}, performs atomic fact verification (measuring recall, precision, and contradiction detection) against expert-curated ground truth rather than Wikipedia or retrieved documents.

Experiments across multiple model configurations on crops and queries from Bihar, India show that fine-tuning on curated data substantially improves fact recall and F1, while maintaining high relevance. Using a fine-tuned smaller model achieves comparable or better factual quality at a fraction of the cost of frontier models. A stitching layer further improves safety subscores while maintaining high conversational quality. We release the \texttt{farmerchat-prompts} library to enable reproducible development of domain-specific agricultural AI.

\end{abstract}


% Main sections
% Introduction
\section{Introduction}
\label{sec:introduction}

The global agricultural sector faces a persistent challenge: providing timely, accurate advice to the over 500 million smallholder farmers who produce roughly one-third of the world's food supply \citep{fao2022digital}. This knowledge gap is especially pronounced in South Asia and Sub-Saharan Africa, where climate variability, fragmented landholdings, and limited access to formal extension services compound the challenge of translating agricultural science into on-farm practice. Traditional extension services remain under-resourced, with extension worker-to-farmer ratios exceeding 1:1000 in many developing regions \citep{aker2011mobilephones}. Mobile-based advisory platforms have emerged as a promising channel to bridge this gap, with systems like \farmerchat{} serving over 1.5 million queries across four countries \citep{sharma2024farmerchat}, demonstrating both the demand for and feasibility of AI-mediated extension at scale. LLMs offer a potential solution, but deploying them for agricultural advisory reveals three systematic limitations. Recent benchmarking confirms these are not merely theoretical concerns: even frontier models achieve below 70\% factuality on general knowledge tasks \citep{deepmind2025factsbenchmark}, underscoring the challenge for specialized domains:

\begin{enumerate}[leftmargin=*]
    \item \textbf{Hallucination risk:} LLMs generate fabricated recommendations that appear credible \citep{ji2023hallucination}, which in agriculture can lead to adverse outcomes from incorrect pesticide dosages, planting times, or disease identification.

    \item \textbf{Lack of specificity:} Models default to generic advice (``apply fertilizer appropriately'') rather than actionable instructions (``apply 120~kg of Urea per hectare at 21 and 45 days after transplanting'').

    \item \textbf{Tone mismatch:} Effective extension requires a warm, culturally appropriate persona \citep{gandhi2009digitalgreen}; generic models produce overly formal responses that fail to build trust.
\end{enumerate}

These challenges motivate our central research questions:

\begin{enumerate}[label=\textbf{RQ\arabic*:},leftmargin=*]
    \item How can we ensure factual accuracy in agricultural LLM responses while minimizing hallucination?
    \item How should domain-specific outputs be evaluated beyond holistic scoring to capture safety-critical errors?
    \item Can fine-tuning with curated agricultural facts improve retrieval without sacrificing conversational quality?
\end{enumerate}

\subsection{Contributions}

This work makes four contributions toward reliable, domain-specific LLM deployment:

\begin{itemize}[leftmargin=*]
    \item \textbf{Hybrid Architecture:} We introduce a two-stage pipeline that decouples factual knowledge retrieval from conversational delivery, enabling independent optimization of accuracy (via LoRA fine-tuning on expert-curated \goldenfacts{}) and communication quality (via prompt-based stitching).

    \item \textbf{Evaluation Framework:} We develop \dgeval{}, a three-level evaluation hierarchy that fills a gap in existing frameworks by performing atomic fact verification against expert-curated ground truth---rather than Wikipedia or retrieved documents---with contradiction detection for safety-critical domains.

    \item \textbf{Empirical Results:} We demonstrate that domain-specific fine-tuning improves fact recall from 26.2\% to 50.3\% and F1 from 37.2\% to 51.8\% (GPT-4o Mini), while enabling 85\% cost reduction by using a fine-tuned smaller model instead of GPT-4.

    \item \textbf{Open-Source Resources:} We release \texttt{farmerchat-prompts}, containing our fact extraction, evaluation, and stitching prompts, along with two public datasets: (1)~a human-curated agricultural QA dataset of expert-verified query--answer pairs,\footnote{\url{https://huggingface.co/datasets/DigiGreen/human_curated_qa_dataset}} and (2)~a human preference evaluation dataset capturing expert pairwise judgments,\footnote{\url{https://huggingface.co/datasets/DigiGreen/human_preference_eval_dataset}} to enable reproducibility and extension to other advisory domains.
\end{itemize}

The remainder of this paper is organized as follows. Section~\ref{sec:related_works} motivates the problem and reviews related work in agricultural AI, LLM fine-tuning, and evaluation frameworks. Section~\ref{sec:human_curation} details the data curation pipeline covering both human expert and synthetic data sources. Section~\ref{sec:methodology} presents our hybrid architecture, data curation pipeline, and the \dgeval{} evaluation framework. Section~\ref{sec:experiments} reports experimental results including benchmark comparisons, ablation studies, and human evaluation. Section~\ref{sec:discussion} discusses key findings, limitations, and future work. Section~\ref{sec:conclusion} concludes.

% Background and Related Work (merged)
\section{Background and Related Work}
\label{sec:related_works}

\subsection{Motivation}
\label{subsec:motivation}

Agricultural advisory differs from general-purpose conversational AI in requiring precise, actionable instructions (specific formulations, dosages, timing), carrying direct safety consequences for incorrect advice, and depending on intersecting contextual variables (crop variety, growth stage, soil, climate). The scale of the challenge is substantial: over 500 million smallholder farmers worldwide depend on timely, accurate agronomic guidance, yet formal extension services remain severely under-resourced across South Asia and Sub-Saharan Africa.

General-purpose LLMs are ill-suited for this domain. Hallucination risk is acute when incorrect dosages or banned pesticide recommendations can cause direct economic harm or health consequences. Models lack the regional specificity needed for actionable advice---locally approved formulations, region-specific planting calendars, and soil-appropriate practices vary substantially even within a single country. Existing RAG-based systems improve grounding but depend on inference-time retrieval infrastructure and assume a comprehensive, well-maintained corpus, which may not hold for specialized agricultural knowledge across diverse geographies.

Equally critical is the evaluation gap: existing factual evaluation frameworks verify against Wikipedia \citep{min2023factscore}, web search \citep{wei2024longform}, or retrieved context \citep{es2024ragas}, none of which capture the expert-curated, safety-critical ground truth required for specialized agricultural domains. Without evaluation frameworks grounded in verified domain knowledge, it is impossible to systematically measure or improve model reliability for high-stakes advisory.

These challenges motivate our hybrid architecture---decoupling verified factual retrieval from conversational delivery---and our domain-specific evaluation framework grounded in expert-curated atomic facts. Our work builds on four research areas.

\subsection{Related Work}
\label{subsec:related_work}

\paragraph{Agricultural AI Systems.}
Although technology-mediated extension has improved farmer outcomes \citep{gandhi2009digitalgreen,cole2021mobile}, information quality in digital agricultural advisory remains a persistent gap. The production \farmerchat{} system has served over 1.5 million queries across India, Kenya, Ethiopia, and Nigeria \citep{sharma2024farmerchat}, demonstrating scalable AI-mediated extension but also revealing systematic accuracy limitations when models operate outside their training distribution. These challenges are not unique to agriculture: across high-stakes advisory domains, the gap between general-purpose LLM capabilities and domain-specific accuracy requirements motivates specialized fine-tuning and evaluation approaches.

\paragraph{LLM Fine-Tuning for Specialized Domains.}
Parameter-efficient fine-tuning via LoRA \citep{hu2022lora} and QLoRA \citep{dettmers2023qlora} has proven effective in high-stakes domains \citep{singhal2023medpalm,chalkidis2020legalbert,wu2023bloomberggpt}. Recent work on hidden factual knowledge suggests that LLMs encode approximately 40\% more knowledge internally than they externalize in responses \citep{gekhman2025insideout}, suggesting that fine-tuning can ``unlock'' latent parametric knowledge rather than merely injecting new information. At the mechanistic level, prior work shows that MLP layers can store facts with optimal parameter efficiency \citep{dugan2025factstoring}, providing a theoretical basis for why targeting specific projection matrices via LoRA is effective for knowledge-intensive tasks. We extend this approach to agriculture, where high factual specificity and safety implications present distinct challenges.

\paragraph{Evaluation Frameworks for Factual Accuracy.}
Existing frameworks address different aspects of factual evaluation: FActScore \citep{min2023factscore} decomposes responses into atomic facts but verifies against Wikipedia; RAGAS \citep{es2024ragas} measures faithfulness to retrieved context; TruthfulQA \citep{lin2022truthfulqa} targets general truthfulness; and LLM-as-a-Judge approaches \citep{zheng2023judging,liu2023geval} correlate with human preferences but lack domain-specific ground truth. More recently, SAFE \citep{wei2024longform} decomposes long-form responses into atomic facts and verifies each via multi-step web search, achieving 72\% agreement with human annotators at 20$\times$ lower cost; however, its reliance on web search limits applicability to specialized domains where authoritative knowledge is not publicly indexed. The FACTS Grounding Leaderboard \citep{cheng2025facts} validates the need for multi-dimensional factuality assessment, evaluating models across four complementary dimensions of grounding quality. None of these approaches combine atomic decomposition with expert-curated verification and safety-critical contradiction detection, which is essential for domains like agriculture where a single incorrect dosage can cause direct harm. \dgeval{} addresses this by decomposing responses into atomic facts verified against expert-curated \goldenfacts{}, with explicit contradiction detection for safety-critical claims.

\paragraph{Hybrid Architectures and Design Motivation.}
RAG \citep{lewis2020rag} combines parametric and non-parametric knowledge but assumes a corpus containing relevant information, which may not hold for specialized agricultural knowledge. The production Farmer.Chat system \citep{sharma2024farmerchat} uses RAG over curated extension documents. RAG and fine-tuning address complementary aspects of knowledge-intensive generation: RAG excels when a well-maintained corpus is available, while fine-tuning can internalize knowledge as parametric recall without inference-time retrieval infrastructure. The present work explores the fine-tuning path, applying a separate stitching layer for conversational adaptation. This decoupling avoids inference-time retrieval dependencies while enabling independent optimization of accuracy and communication style; a direct comparison of the two approaches remains important future work.

\begin{figure}[t]
\centering
\includegraphics[width=0.7\textwidth]{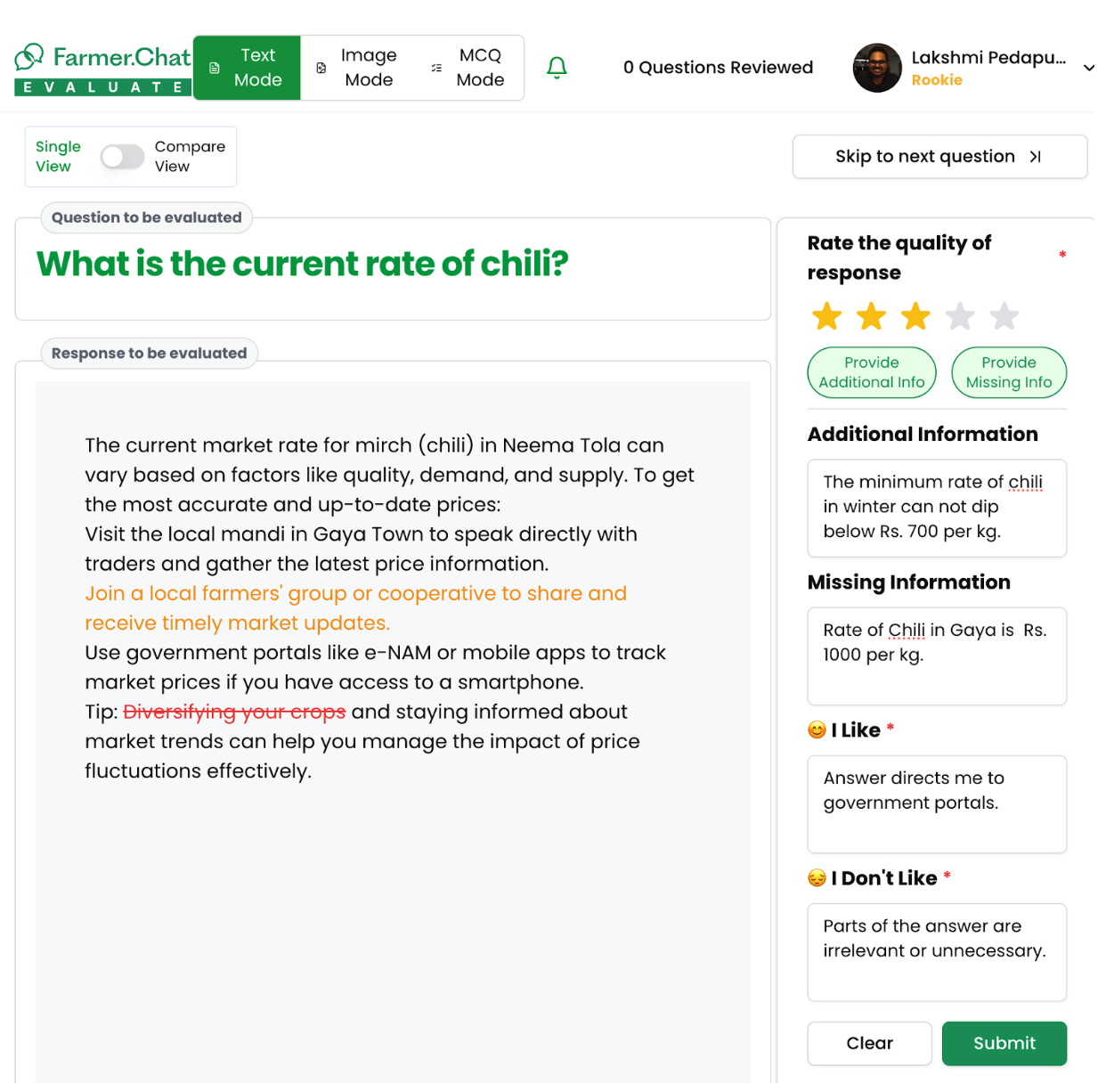}
\caption{The \texttt{evaluate.farmer.chat} platform used for expert curation and evaluation. The interface supports quality rating, additional/missing information annotation, like/dislike feedback, and side-by-side comparison of model responses, enabling systematic human evaluation of agricultural advisory content. Specific user inputs shown in the image may not be accurate and are for illustration purposes only.}
\label{fig:farmer_chat_evaluate}
\end{figure}

% Data Curation Pipeline
\section{Data Curation Pipeline}
\label{sec:human_curation}

Our training data derives from two complementary sources: human-curated Golden Answers produced by domain experts through the \texttt{evaluate.farmer.chat} platform (Section~\ref{subsec:human_expert_curation}), and synthetically enriched data generated through multi-source augmentation (Section~\ref{subsec:synthetic_curation}). Both sources undergo the same \goldenfact{} extraction pipeline (Section~\ref{subsec:data_curation}) to produce atomic, verified training signals. Section~\ref{subsec:ablation} examines how the balance between these sources affects downstream performance.

\subsection{Human Expert Data Curation Pipeline}
\label{subsec:human_expert_curation}

Large language models trained on broad internet corpora lack the localized, safety-critical agricultural knowledge required for reliable farmer advisory. To address this, we developed a systematic human-in-the-loop curation pipeline that produces expert-verified training data at scale. Over 25,000 query--answer pairs have been reviewed across India, Kenya, Ethiopia, and Nigeria; 11,966 validated pairs are used in this work.\footnote{The curated QA dataset is publicly available at \url{https://huggingface.co/datasets/DigiGreen/human_curated_qa_dataset}.} The cross-country deployment revealed important regional adaptation patterns: while core agronomic principles (e.g., nutrient management ratios) transfer across geographies, pest and disease pressures, locally approved pesticide formulations, and cropping calendars require region-specific expert input. For example, rice pest management recommendations differ substantially between Bihar (India) and Western Kenya due to distinct pest complexes and regulatory environments. A companion paper \citep{singh2026humancuration} explores preference-based alignment using the evaluation data from this curation pipeline.

\subsection{Annotation Platform and Evaluation Modes}

The curation pipeline is built on the \texttt{evaluate.farmer.chat} platform (Figure~\ref{fig:farmer_chat_evaluate}), a web-based annotation tool supporting two complementary evaluation modes:

\textbf{Absolute evaluation.} Reviewers score each response on a 5-point Likert scale across five dimensions: factual accuracy, clarity, tone appropriateness, safety, and contextual relevance. Reviewers also provide suggested edits, and these corrected responses become the \textbf{Golden Answers} used for training. The correction workflow is structured: reviewers assess each LLM response against their domain expertise, marking specific spans that require correction---factual errors, missing details, or inappropriate advice---then provide a revised response. Reviewers augment responses with region-specific information such as locally approved pesticide formulations, Bihar-specific planting calendars, and soil-appropriate nutrient management practices. Each Golden Answer therefore reflects expert synthesis rather than simple acceptance or rejection of the original model output.

\textbf{Comparative evaluation.} Reviewers select a preferred response from side-by-side model variants, providing written justification. This preference data informed iterative prompt refinement during development; its use for preference-based alignment is explored in the companion paper \citep{singh2026humancuration}.

\subsection{Query Selection and Reviewer Methodology}

Queries are prioritized for review based on four criteria: frequency (high-volume query clusters), novelty or ambiguity (underrepresented topics), model confidence (low-confidence responses flagged for expert attention), and prior reviewer flags from earlier review rounds.

The review team comprises four domain experts with 5--10 years of experience spanning agricultural extension officers, NGO agronomists, and lead farmer trainers. Reviewers are based in Bihar with additional remote participation, and are bilingual in Hindi and English. Onboarding includes structured workshops covering scoring rubrics, platform usage, and calibration exercises. Reviewers are compensated fairly for their time.

\subsection{Quality Assurance and Safety}

Multiple mechanisms ensure annotation consistency and safety:

\begin{itemize}[leftmargin=*]
    \item \textbf{Double review:} 20\% of pairs are independently scored by two reviewers to measure inter-rater reliability. Overlap pairs are selected via stratified sampling across crop types and query categories to ensure representative coverage. Agreement is measured at both the response level (overall score concordance) and the dimension level (per-dimension score alignment). Disagreements exceeding one point on any dimension are escalated to adjudication by the senior agronomist.
    \item \textbf{Control pairs:} 10\% of the review queue consists of previously scored responses, enabling detection of scorer drift over time.
    \item \textbf{Adjudication:} A senior agronomist resolves disagreements on contested topics (e.g., conflicting regional practices).
    \item \textbf{Calibration sessions:} Periodic group review sessions maintain scoring alignment across the team.
    \item \textbf{Red-line protocol:} Responses recommending banned pesticides, omitting required safety instructions, or containing gender-insensitive content are flagged and excluded.
    \item \textbf{Multiple valid answers:} Where legitimate alternatives exist (e.g., organic vs.\ chemical approaches), both are flagged rather than forcing a single ``correct'' answer.
\end{itemize}

Inter-reviewer agreement patterns vary systematically by content type: quantitative recommendations (dosages, application rates) show high agreement ($>$90\% concordance), while practice-dependent advice (organic vs.\ chemical approaches, timing relative to local conditions) shows more variability, reflecting legitimate regional and methodological differences rather than annotation noise.

\noindent Full audit trails are maintained for all scoring decisions. The validated Golden Answers produced by this pipeline serve as input to the \goldenfact{} extraction process described in Section~\ref{subsec:data_curation}.

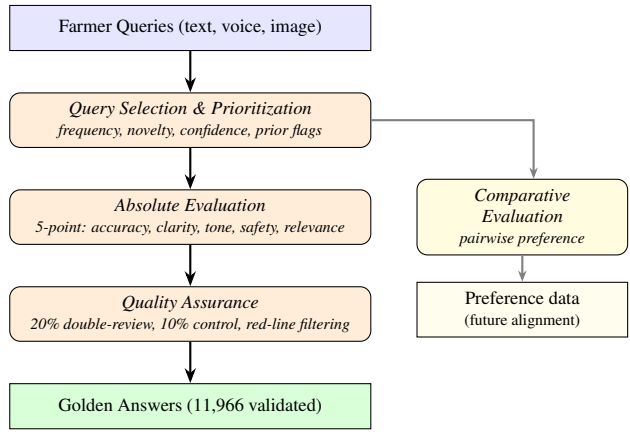
\begin{figure}[t]
\centering
\begin{tikzpicture}[
    node distance=0.55cm,
    data/.style={rectangle, draw, fill=blue!10, minimum height=0.6cm,
                 minimum width=4.8cm, align=center, font=\scriptsize},
    process/.style={rectangle, rounded corners, draw, fill=orange!15,
                    minimum height=0.6cm, minimum width=4.8cm, align=center,
                    font=\scriptsize\itshape},
    sideprocess/.style={rectangle, rounded corners, draw, fill=yellow!15,
                    minimum height=0.6cm, minimum width=2.8cm, align=center,
                    font=\scriptsize\itshape},
    arrow/.style={-{Stealth[length=2mm]}, thick},
    sidearrow/.style={-{Stealth[length=1.5mm]}, thick, gray}
]

% Main vertical flow
\node[data] (queries) {Farmer Queries (text, voice, image)};
\node[process, below=of queries] (selection) {Query Selection \& Prioritization\\{\tiny frequency, novelty, confidence, prior flags}};
\node[process, below=of selection] (absolute) {Absolute Evaluation\\{\tiny 5-point: accuracy, clarity, tone, safety, relevance}};
\node[process, below=of absolute] (qa) {Quality Assurance\\{\tiny 20\% double-review, 10\% control, red-line filtering}};
\node[data, below=of qa, fill=green!15] (golden) {Golden Answers (11,966 validated)};

% Side branch: comparative evaluation
\node[sideprocess, right=0.6cm of absolute] (comparative) {Comparative\\Evaluation\\{\tiny pairwise preference}};
\node[data, below=0.35cm of comparative, fill=yellow!8, minimum width=2.8cm, font=\scriptsize] (prefdata) {Preference data\\{\tiny (future alignment)}};

% Arrows
\draw[arrow] (queries) -- (selection);
\draw[arrow] (selection) -- (absolute);
\draw[arrow] (absolute) -- (qa);
\draw[arrow] (qa) -- (golden);
\draw[sidearrow] (selection.east) -| ([xshift=0.15cm]comparative.north);
\draw[sidearrow] (comparative) -- (prefdata);

\end{tikzpicture}
\caption{Expert data curation pipeline. Farmer queries are prioritized by frequency and model confidence, then reviewed by domain experts using absolute scoring across five dimensions. Quality assurance includes partial double-review and control pairs for consistency tracking. Comparative evaluation produces preference data for future alignment work.}
\label{fig:curation_pipeline}
\end{figure}

\subsection{Synthetic Data Curation Pipeline}
\label{subsec:synthetic_curation}

To extend coverage beyond the 11,966 human-curated pairs, we generate synthetic training data from multiple complementary sources. This augmentation targets underrepresented crops, rare pest scenarios, and emerging agronomic practices where expert-curated data is sparse.

\paragraph{Data sources.} Synthetic data is drawn from five sources:
\begin{enumerate}[leftmargin=*]
    \item \textbf{Document RAG:} Retrieval-augmented generation over Digital Green's curated extension documents, agricultural manuals, and government advisory bulletins.
    \item \textbf{Video RAG:} Extraction and synthesis from transcripts of agricultural extension videos, capturing practical demonstrations and farmer-facing explanations.
    \item \textbf{LLM synthesis:} Direct generation using GPT-4o and GPT-4.1, prompted with crop-specific templates and regional context to produce query--answer pairs for underrepresented topics.
    \item \textbf{Web search augmentation:} Targeted web retrieval for emerging practices, recent cultivar recommendations, and updated regulatory information not yet reflected in static documents.
    \item \textbf{Cross-source synthesis:} Integration across multiple sources for complex queries requiring information from several documents or modalities.
\end{enumerate}

\paragraph{Golden Fact extraction.} All synthetic responses undergo the same three-step \goldenfact{} extraction pipeline applied to human-curated data: semantic grouping to merge equivalent recommendations, contradiction detection to identify conflicting claims, and finalization to produce minimal atomic statements (Section~\ref{subsec:data_curation}).

\paragraph{Quality scoring.} Each extracted fact is scored along three dimensions: \emph{confidence} (alignment with authoritative sources), \emph{completeness} (presence of necessary parameters such as dosage, timing, and method), and \emph{actionability} (whether a farmer could implement the recommendation directly). Facts scoring below threshold on any dimension are excluded from the training set.

\paragraph{Source traceability.} Every synthetic fact retains provenance metadata linking it to its originating source(s), enabling downstream auditing and targeted re-curation when source documents are updated.

\paragraph{Quality--quantity trade-off.} The ablation study in Section~\ref{subsec:ablation} examines how the balance between human-curated and synthetic data affects model performance, revealing a quality--quantity threshold effect: 12k human-curated facts outperform 62k mixed-source facts, but scaling synthetic data to 130k achieves the best overall precision, suggesting that sufficient synthetic volume can compensate for lower per-example quality.

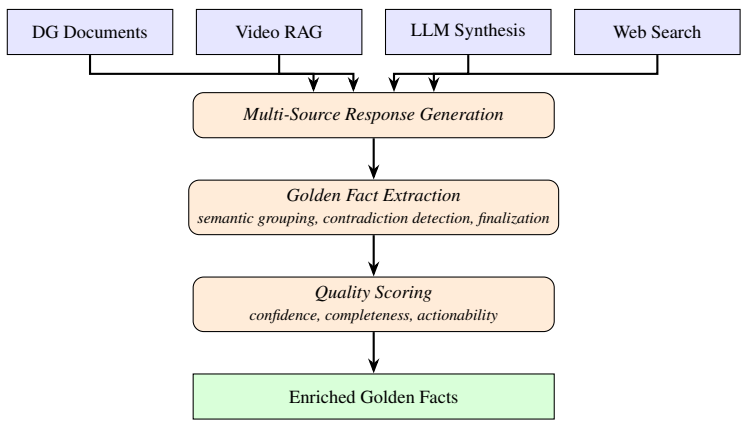
\begin{figure}[t]
\centering
\begin{tikzpicture}[
    node distance=0.55cm and 0.4cm,
    source/.style={rectangle, draw, fill=blue!10, minimum height=0.6cm,
                   minimum width=2.2cm, align=center, font=\scriptsize},
    process/.style={rectangle, rounded corners, draw, fill=orange!15,
                    minimum height=0.6cm, minimum width=4.8cm, align=center,
                    font=\scriptsize\itshape},
    data/.style={rectangle, draw, fill=green!15, minimum height=0.6cm,
                 minimum width=4.8cm, align=center, font=\scriptsize},
    arrow/.style={-{Stealth[length=2mm]}, thick}
]

% Top row: 4 source nodes
\node[source] (docs) {DG Documents};
\node[source, right=0.3cm of docs] (video) {Video RAG};
\node[source, right=0.3cm of video] (llm) {LLM Synthesis};
\node[source, right=0.3cm of llm] (web) {Web Search};

% Flow nodes
\node[process, below=0.8cm of $(video)!0.5!(llm)$] (generation) {Multi-Source Response Generation};
\node[process, below=of generation] (extraction) {Golden Fact Extraction\\{\tiny semantic grouping, contradiction detection, finalization}};
\node[process, below=of extraction] (scoring) {Quality Scoring\\{\tiny confidence, completeness, actionability}};
\node[data, below=of scoring] (facts) {Enriched Golden Facts};

% Arrows from sources to generation
\draw[arrow] (docs.south) -- ++(0,-0.25) -| ([xshift=-0.8cm]generation.north);
\draw[arrow] (video.south) -- ++(0,-0.25) -| ([xshift=-0.27cm]generation.north);
\draw[arrow] (llm.south) -- ++(0,-0.25) -| ([xshift=0.27cm]generation.north);
\draw[arrow] (web.south) -- ++(0,-0.25) -| ([xshift=0.8cm]generation.north);

% Arrows between process nodes
\draw[arrow] (generation) -- (extraction);
\draw[arrow] (extraction) -- (scoring);
\draw[arrow] (scoring) -- (facts);

\end{tikzpicture}
\caption{Synthetic data curation pipeline. Four complementary sources feed into multi-source response generation. All synthetic responses undergo the same Golden Fact extraction pipeline as human-curated data, followed by quality scoring to filter low-confidence or incomplete facts.}
\label{fig:synthetic_pipeline}
\end{figure}

% Methodology
\section{Methodology}
\label{sec:methodology}

Our methodology addresses five design principles, detailed below through data curation (Section~\ref{subsec:data_curation}), the hybrid architecture (Section~\ref{subsec:hybrid_engine}), and evaluation (Section~\ref{subsec:evaluation_framework}): (1)~expert-verified facts as the training signal, (2)~atomic claim decomposition for precise evaluation, (3)~systematic safety coverage including contradiction detection, (4)~independent optimization of accuracy and conversational quality, and (5)~efficient model targeting for cost-effective deployment.

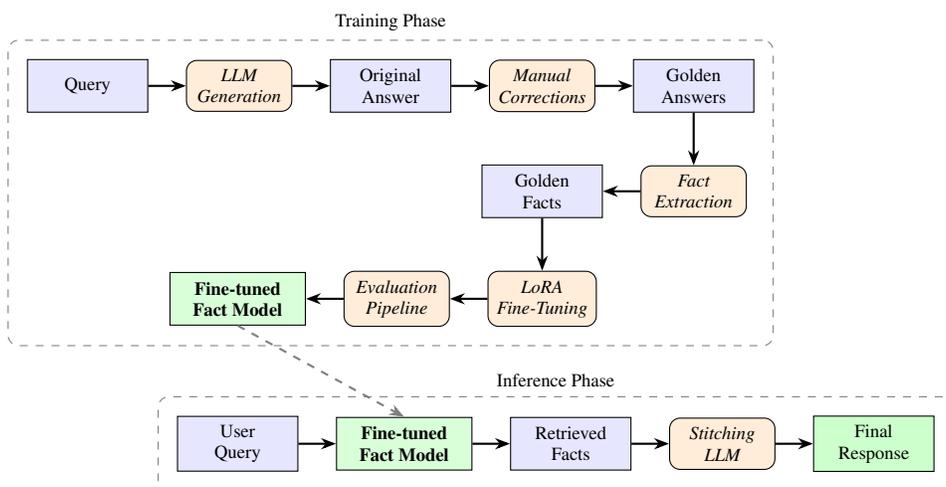
\begin{figure}[!t]
\centering
\begin{tikzpicture}[
    node distance=0.7cm and 0.5cm,
    data/.style={rectangle, draw, fill=blue!10, minimum height=0.7cm,
                 minimum width=1.6cm, align=center, font=\scriptsize},
    process/.style={rectangle, rounded corners, draw, fill=orange!15,
                    minimum height=0.6cm, minimum width=1.4cm, align=center, font=\scriptsize\itshape},
    model/.style={rectangle, draw, fill=green!15, minimum height=0.7cm,
                  minimum width=1.8cm, align=center, font=\scriptsize\bfseries},
    arrow/.style={-{Stealth[length=2mm]}, thick},
    phase/.style={draw=gray, dashed, rounded corners, inner sep=0.25cm}
]

% TRAINING PHASE (top)
% Row 1: Query → LLM Gen → Original Answer → Manual Corrections → Golden Answers
\node[data] (query) {Query};
\node[process, right=of query] (llmgen) {LLM\\Generation};
\node[data, right=of llmgen] (original) {Original\\Answer};
\node[process, right=of original] (manual) {Manual\\Corrections};
\node[data, right=of manual] (golden) {Golden\\Answers};

% Row 2: Golden Facts ← Fact Extraction
\node[process, below=of golden] (extract) {Fact\\Extraction};
\node[data, left=of extract] (facts) {Golden\\Facts};

% Row 3: Fine-tuning → Evaluation → Model
\node[process, below=of facts] (finetune) {LoRA\\Fine-Tuning};
\node[process, left=of finetune] (eval) {Evaluation\\Pipeline};
\node[model, left=of eval] (model) {Fine-tuned\\Fact Model};

% Training arrows
\draw[arrow] (query) -- (llmgen);
\draw[arrow] (llmgen) -- (original);
\draw[arrow] (original) -- (manual);
\draw[arrow] (manual) -- (golden);
\draw[arrow] (golden) -- (extract);
\draw[arrow] (extract) -- (facts);
\draw[arrow] (facts) -- (finetune);
\draw[arrow] (finetune) -- (eval);
\draw[arrow] (eval) -- (model);

% INFERENCE PHASE (bottom) - positioned below training phase
\node[data, below=1.2cm of model] (userq) {User\\Query};
\node[model, right=of userq] (ftmodel) {Fine-tuned\\Fact Model};
\node[data, right=of ftmodel] (retfacts) {Retrieved\\Facts};
\node[process, right=of retfacts] (stitch) {Stitching\\LLM};
\node[data, right=of stitch, fill=green!20] (response) {Final\\Response};

% Inference arrows
\draw[arrow] (userq) -- (ftmodel);
\draw[arrow] (ftmodel) -- (retfacts);
\draw[arrow] (retfacts) -- (stitch);
\draw[arrow] (stitch) -- (response);

% Connection from training to inference
\draw[arrow, dashed, gray] (model.south) -- (ftmodel.north);

% Phase labels
\begin{scope}[on background layer]
    \node[phase, fit=(query)(golden)(model)(finetune), label={[font=\scriptsize]above:Training Phase}] {};
    \node[phase, fit=(userq)(response), label={[font=\scriptsize]above:Inference Phase}] {};
\end{scope}

\end{tikzpicture}
\caption{Hybrid Engine Architecture. \textit{Top:} Training pipeline showing data curation and LoRA fine-tuning. \textit{Bottom:} Inference pipeline where the fine-tuned model retrieves \goldenfacts{}, which are transformed into conversational responses by the stitching layer.}
\label{fig:architecture}
\end{figure}

\begin{table}[t]
\centering
\caption{Dataset Statistics}
\label{tab:dataset_stats}
\begin{tabular}{lc}
\toprule
\textbf{Statistic} & \textbf{Value} \\
\midrule
Total unique queries & 11,966 \\
Expert-validated Golden Answers & 11,966 \\
Extracted \goldenfacts{} & 110,723 \\
Average facts per answer & 9.23 \\
Crops covered & 60 \\
Languages & English and Hindi \\
Geographic regions & Bihar, India \\
\bottomrule
\end{tabular}
\end{table}

\subsection{Data Curation: From Queries to Golden Facts}
\label{subsec:data_curation}

Our training data originates from the human expert curation pipeline detailed in Section~\ref{sec:human_curation}. In brief: real-world farmer queries from the \farmerchat{} platform are clustered by embedding similarity \citep{reimers2019sentencebert}, prioritizing the ${\sim}$20\% of query types that account for 70\% of volume. Domain experts review clustered queries through the \texttt{evaluate.farmer.chat} platform (Figure~\ref{fig:farmer_chat_evaluate}) to produce \textbf{Golden Answers}---verified responses serving as ground truth. Coverage is extended through the synthetic data curation pipeline (Section~\ref{subsec:synthetic_curation}), with all sources undergoing the same \goldenfact{} extraction process.

Golden Answers are too complex for direct model training. We decompose them into \textbf{\goldenfacts{}}: atomic, non-conflicting, actionable, unique, and complete units of agricultural information, each containing the minimal details (dosage, timing, method) needed for a single implementable recommendation. For example: ``Apply first Urea dose of 60~kg per hectare at 21 days after transplanting rice.'' A compound recommendation such as ``Apply 120~kg of Urea per hectare in two split doses at 21 and 45 days after transplanting'' would be decomposed into separate facts for dosage, schedule, and timing (see Appendix~\ref{app:prompts}).

\goldenfacts{} are extracted from Golden Answers via a three-step prompt pipeline: \textbf{(1)~Semantic grouping} merges equivalent recommendations that differ only in surface phrasing, reducing the fact set by ${\sim}$15\%. \textbf{(2)~Contradiction detection} identifies conflicting dosages, timing, or polarities, classified by severity; high-severity contradictions trigger expert re-review. \textbf{(3)~Finalization} produces minimal atomic statements, each self-contained with exactly one actionable recommendation and its necessary parameters.

\subsection{The Hybrid Engine: Fine-Tuning and Fact Stitching}
\label{subsec:hybrid_engine}

Our hybrid architecture decouples knowledge retrieval from conversational delivery through a two-stage pipeline (Figure~\ref{fig:architecture}).

\subsubsection{Stage 1: Fine-Tuning for Fact Retrieval}

The first stage optimizes the LLM for \textbf{structured fact retrieval}. Rather than training for conversational quality, we use Supervised Fine-Tuning (SFT) with Low-Rank Adaptation (LoRA) \citep{hu2022lora} on query-to-facts pairs derived from our curated dataset.

\textbf{Training Objective:} Given a farmer query $q$, the model is trained to output the set of relevant \goldenfacts{} $\{f_1, f_2, \ldots, f_n\}$ in a structured, parseable format, prioritizing high recall of the golden set and specificity (precise values rather than hedged ranges).

\textbf{Model Selection and Configuration:} We focus on efficient models (GPT-4o-mini, Llama 3 8B) suitable for cost-effective deployment, fine-tuned with LoRA \citep{hu2022lora}. Our LoRA configuration uses rank $r{=}8$ with scaling factor $\alpha{=}16$ (effective scaling $\alpha/r = 2$), targeting query, key, and value projection matrices in all attention layers. We train for 3 epochs with a cosine learning rate schedule (peak $\text{lr} = 2 \times 10^{-4}$, 10\% warmup), batch size 16, and LoRA dropout 0.05. Mechanistic evidence indicates that MLP layers are primary sites of factual knowledge storage \citep{dugan2025factstoring}; targeting QKV projections alongside MLPs is standard LoRA practice \citep{hu2022lora} and provides efficient coverage for domain-specific knowledge injection. Fine-tuning smaller models may reduce reliance on potentially unreliable parametric knowledge, encouraging the model to draw more heavily on fine-tuned facts.

\subsubsection{Stage 2: The Stitching Layer}

Once accurate facts are retrieved, a second LLM call applies the \textbf{Fact Stitching Prompt}. This layer transforms dry factual output into natural, farmer-appropriate responses.

The stitching layer implements the \farmerchat{} persona---culturally appropriate language, automatic safety precautions, and structured organization---without altering the underlying agronomy (see Appendix~\ref{app:prompts} for prompt details).

\subsection{Evaluation Framework: DG-Eval}
\label{subsec:evaluation_framework}

\begin{table}[!t]
\centering
\caption{Summary of \dgeval{} Metrics}
\label{tab:eval_metrics}
\small
\begin{tabularx}{\textwidth}{p{0.16\textwidth} p{0.22\textwidth} p{0.22\textwidth} X}
\toprule
\textbf{Level} & \textbf{Category} & \textbf{Metrics} & \textbf{Description} \\
\midrule
\multirow{2}{*}{1. Intrinsic} & Specificity & Contextual anchors & Actionable, precise advice \\
& Conversationality & 6-dimension score & Persona adherence \\
\midrule
2. Query & Relevance & Relevance score & Addresses user's question \\
\midrule
\multirow{3}{*}{3. Ground Truth} & Fact Recall & Completeness & Coverage of \goldenfacts{} \\
& Fact Precision & Correctness & Accuracy of generated facts \\
& Contradiction & Safety flags & Identifies dangerous conflicts \\
\bottomrule
\end{tabularx}
\end{table}

\begin{figure}[!t]
\centering
\begin{tikzpicture}[
    box/.style={draw=black!40, rounded corners=3pt, text width=11.2cm,
                inner sep=5pt, outer sep=0pt, font=\footnotesize},
    legendbox/.style={draw=black!40, rounded corners=3pt, fill=gray!5,
                      inner sep=6pt, font=\footnotesize, anchor=north west}
]

%% ---------- left column: Q / A1 / A2 ----------

% Question
\node[box, fill=gray!8] (Q) {%
    \textbf{Q:} How can soil health be improved for better chili yields?};

% Answer 1 — low specificity (3/7)
\node[box, fill=white, below=6pt of Q] (A1) {%
    \textbf{A1 \normalfont(Low specificity --- 3/7):}\par\smallskip
    \hlc{cAction}{Use} \hlc{cEntity}{liquid jeevamritam} to improve
    the soil. It should be applied \hlc{cQuantity}{3 to 4 times}
    during the crop season. This helps improve soil fertility and
    can lead to better yields.};

% Answer 2 — high specificity (7/7)
\node[box, fill=white, below=6pt of A1] (A2) {%
    \textbf{A2 \normalfont(High specificity --- 7/7):}\par\smallskip
    \hlc{cAction}{Apply} \hlc{cQuantity}{200 liters per acre} of
    \hlc{cEntity}{Jeevamrit} to your \hlc{cEntity}{chili} field
    \hlc{cTime}{every 20 days} during the
    \hlc{cTime}{crop season (June--October)}.
    This will
    \hlc{cMechanism}{enrich the soil with beneficial microbes,}
    \hlc{cMechanism}{improving nutrient availability}
    \hlc{cCompare}{more effectively than}
    \hlc{cCompare}{chemical fertilizers alone},
    especially on \hlc{cLocation}{clay-loam soils of Patna, Bihar}.};

%% ---------- legend (right of the Q/A boxes, with bounding box) ----------
\node[legendbox] (legend) at ([xshift=8pt]Q.north east) {%
    \begin{tabular}{@{}c@{\,}l@{}}
    \multicolumn{2}{@{}l@{}}{\textbf{Legend}}\\[4pt]
    \hline\noalign{\vskip 2pt}
    \colorbox{cAction}{\rule{0.45cm}{0pt}\rule{0pt}{0.22cm}}
        & \hlc{cAction}{Actionable}\\[2pt]
    \hline\noalign{\vskip 2pt}
    \colorbox{cEntity}{\rule{0.45cm}{0pt}\rule{0pt}{0.22cm}}
        & \hlc{cEntity}{Entity}\\[2pt]
    \hline\noalign{\vskip 2pt}
    \colorbox{cLocation}{\rule{0.45cm}{0pt}\rule{0pt}{0.22cm}}
        & \hlc{cLocation}{Location}\\[2pt]
    \hline\noalign{\vskip 2pt}
    \colorbox{cTime}{\rule{0.45cm}{0pt}\rule{0pt}{0.22cm}}
        & \hlc{cTime}{Time}\\[2pt]
    \hline\noalign{\vskip 2pt}
    \colorbox{cQuantity}{\rule{0.45cm}{0pt}\rule{0pt}{0.22cm}}
        & \hlc{cQuantity}{Quantity}\\[2pt]
    \hline\noalign{\vskip 2pt}
    \colorbox{cCompare}{\rule{0.45cm}{0pt}\rule{0pt}{0.22cm}}
        & \hlc{cCompare}{Conditional}\\[2pt]
    \hline\noalign{\vskip 2pt}
    \colorbox{cMechanism}{\rule{0.45cm}{0pt}\rule{0pt}{0.22cm}}
        & \hlc{cMechanism}{Mechanistic}\\[2pt]
    \hline
    \end{tabular}};

\end{tikzpicture}
\caption{Specificity illustration. Answer~1 triggers only 3 of 7 contextual anchors (sparse highlighting); Answer~2 triggers all 7 (dense highlighting), immediately conveying the difference between generic and specific agricultural advice.}
\label{fig:specificity}
\end{figure}
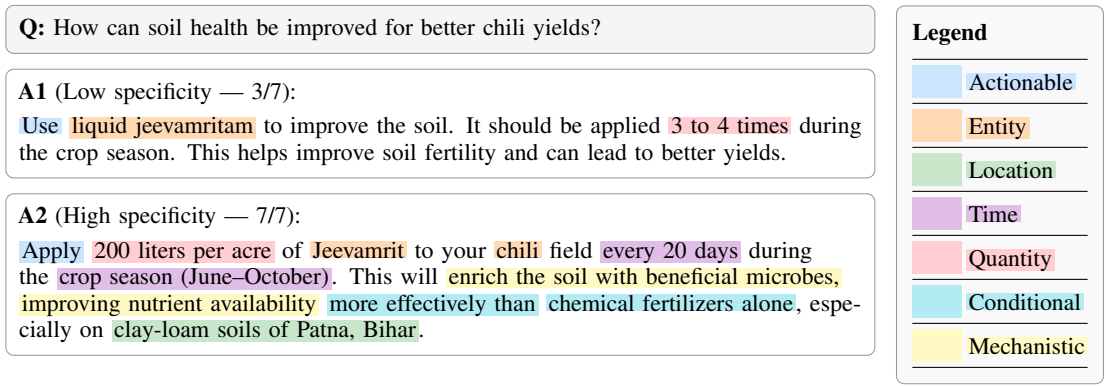

Standard NLG metrics fail to capture the safety-critical requirements of agricultural advisory. We developed \dgeval{}, a multi-dimensional evaluation framework organized into three levels summarized in Table~\ref{tab:eval_metrics}.

\subsubsection{Level 1: Intrinsic Quality}

This level assesses response quality independent of any reference, focusing on two dimensions:

\textbf{Specificity} evaluates whether a response is comprehensive, action-oriented, agriculture-focused, and provides specific actionable advice rather than generic verbose content. We operationalize specificity through seven contextual anchors (Figure~\ref{fig:specificity}) and summarize them below:

\begin{itemize}[leftmargin=*]
    \item \textbf{Actionable:} Clear, implementable instructions (e.g., ``apply,'' ``spray'')
    \item \textbf{Entity-Specific:} Named crops, pests, chemicals, or inputs
    \item \textbf{Location-Specific:} Regional, soil-type, or agro-climatic anchors
    \item \textbf{Time-Specific:} Growth stages, calendar windows, or frequency
    \item \textbf{Quantity-Specific:} Measurable dosages, rates, counts, or areas
    \item \textbf{Conditional/Comparison:} Relative effectiveness or trade-off statements
    \item \textbf{Mechanistic:} Explanation of why a practice works (biological/chemical rationale)
\end{itemize}

\textbf{Conversationality} evaluates adherence to the \farmerchat{} persona using LLM-as-a-Judge scoring \citep{zheng2023judging} across six dimensions, each scored 1--5: content quality (perceived coherence and informativeness, without reference-based accuracy checking), communication style, practical advice feasibility, safety and credibility, conversation flow, and response format. The overall conversationality score is the aggregate across dimensions.

\subsubsection{Level 2: Query Alignment}

A response may be specific and conversational yet fail to address the farmer's actual question. We measure \textbf{Relevance}: the extent to which the answer addresses the core question, avoids tangential information, and provides practical value for the farmer's immediate need.

\subsubsection{Level 3: Ground Truth Alignment}

The most critical level measures factual accuracy against expert-verified \goldenfacts{}. Both the generated response and the Golden Answer are decomposed into atomic facts for comparison. This decomposition-then-verify approach follows a similar strategy to SAFE \citep{wei2024longform}, which decomposes responses into atomic facts verified via web search; our key difference is the use of expert-curated \goldenfacts{} as the verification source rather than web search, which is essential for specialized domains where authoritative knowledge is not publicly indexed. We compute standard precision, recall, and F1 over the overlap between \goldenfacts{} (GF) and retrieved facts (RF): recall measures what fraction of GF appears in RF (completeness), precision measures what fraction of RF appears in GF (correctness), and F1 is their harmonic mean.

\textbf{Contradiction Detection} identifies dangerous conflicts between generated and ground truth facts. High recall and precision are insufficient if the model includes contradictory information. Our contradiction detection prompt compares:

\begin{itemize}[leftmargin=*]
    \item Numeric values (dosage, timing, quantities)
    \item Polarities (do vs. do not)
    \item Method compatibilities (organic vs. chemical approaches)
\end{itemize}

% Experiments
\section{Experiments}
\label{sec:experiments}

\subsection{Experimental Setup}
\label{subsec:setup}

We partition our curated dataset into training (9572 query-fact pairs, 75\%), validation (1197, 12.5\%), and test (1197, 12.5\%) splits, stratified by crop type and query category. Test queries are held out from all model training and prompt development.

We evaluate ten baseline models: GPT-4 \citep{achiam2023gpt4}, GPT-4o and GPT-4o Mini \citep{openai2024gpt4o}, Llama 3 8B \citep{grattafiori2024llama3}, Llama 3.3 70B \citep{meta2024llama33}, Gemma 2 9B and Gemma 2 27B \citep{team2024gemma2}, Gemini 1.5 Pro \citep{team2024gemini15}, Qwen 3 235B \citep{qwen2025qwen3}, and Kimi K2 \citep{moonshot2025kimik2}, alongside fine-tuned variants of GPT-4o, GPT-4o Mini, and Llama 3 8B. All models use the same prompt template and system instructions; however, models with different instruction formats (e.g., Qwen 3, Kimi K2) may be more sensitive to prompt phrasing, which could partially explain performance differences. Fine-tuning uses NVIDIA H100 GPUs (Together AI, OpenAI infrastructure); evaluation uses GPT-4o as judge for conversationality scoring. We note that using GPT-4o as judge while also evaluating GPT-4o variants may introduce self-preference bias \citep{zheng2023judging}; our primary metrics (recall, precision, F1) are ground-truth-based and unaffected, and our human evaluation (Section~\ref{subsec:human_eval}) corroborates the automated rankings.

\subsection{Benchmark Results}
\label{subsec:benchmark}

\subsubsection{Ground Truth Alignment}

Table~\ref{tab:benchmark_results} presents our primary benchmark results including ground truth alignment metrics and overall conversationality scores.\footnote{Paired t-tests confirm significance: GPT-4o Mini F1 improvement from 37.2\% to 51.8\% ($p < 0.001$, $n = 1197$); recall improvement from 26.2\% to 50.3\% ($p < 0.001$); precision decrease from 64.3\% to 53.3\% ($p < 0.01$).}

\begin{table}[t]
\centering
\caption{Benchmark Results: Ground Truth Alignment and Conversationality}
\label{tab:benchmark_results}
\small
\begin{tabular}{lccccc}
\toprule
\textbf{Model} & \textbf{Recall} & \textbf{Precision} & \textbf{F1} & \textbf{Relevance} & \textbf{Conv.} \\
\midrule
\multicolumn{6}{l}{\textit{Baseline Models}} \\
GPT-4 & 26.6\% & 62.2\% & 37.3\% & 96.9\% & 4.01 \\
GPT-4o & 28.5\% & 64.9\% & 39.6\% & 95.8\% & 3.95 \\
GPT-4o Mini & 26.2\% & 64.3\% & 37.2\% & 94.4\% & 3.98 \\
Llama 3 8B & 13.4\% & \textbf{94.2\%} & 23.5\% & \textbf{98.3\%} & 3.72 \\
Llama 3.3 70B & 24.1\% & 58.9\% & 34.2\% & 90.0\% & --- \\
Gemma 2 9B & 32.2\% & 49.6\% & 39.1\% & 90.1\% & 4.21 \\
Gemma 2 27B & 35.7\% & 38.4\% & 37.0\% & 93.2\% & --- \\
Gemini 1.5 Pro & 42.1\% & 56.2\% & 48.2\% & 92.6\% & --- \\
Qwen 3 235B & 19.2\% & 31.3\% & 23.8\% & 55.9\% & --- \\
Kimi K2 & 18.7\% & 27.1\% & 22.1\% & 50.2\% & --- \\
\midrule
\multicolumn{6}{l}{\textit{Fine-Tuned Models}} \\
GPT-4o FT (12k) & 51.1\% & 52.2\% & 51.7\% & 88.8\% & 4.58 \\
GPT-4o Mini FT (12k) & 50.3\% & 53.3\% & 51.8\% & 87.5\% & --- \\
GPT-4o Mini FT (130k) & 43.8\% & 66.2\% & \textbf{52.7\%} & 91.6\% & \textbf{4.62} \\
Llama 3 8B FT (12k) & 15.1\% & 84.3\% & 25.6\% & 91.8\% & 3.89 \\
GPT-4o FT (130k) & \textbf{58.7\%} & 54.9\% & \textbf{56.7\%} & 90.9\% & --- \\
\bottomrule
\end{tabular}

\vspace{2pt}
{\footnotesize All training data configurations are detailed in Table~\ref{tab:ablation_gpt4o}. ``---'' indicates conversationality was not evaluated for this model due to evaluation resource constraints.}
\end{table}

Fine-tuning improves recall by 22--25 percentage points: GPT-4o improves from 28.5\% to 51.1\%, while GPT-4o Mini improves from 26.2\% to 50.3\%. This comes with a precision--recall trade-off (vanilla GPT-4o Mini: 64.3\% precision vs.\ 53.3\% fine-tuned), acceptable since recall matters more for completeness of advice in this domain. The precision--recall shift reflects a behavioral change: baseline models produce conservative, high-precision but sparse responses, while fine-tuned models generate more comprehensive outputs that cover a larger fraction of the golden fact set at the cost of occasionally including marginally relevant information.

Llama 3 8B shows limited improvement (13.4\%$\rightarrow$15.1\%), which may reflect model capacity constraints, though other factors (tokenizer mismatch, hyperparameter sensitivity) could also contribute. Notably, Llama 3 8B's baseline exhibits an unusually sparse response pattern---achieving 94.2\% precision by generating very few facts per response, effectively prioritizing safety over completeness. This conservative behavior persists after fine-tuning, suggesting that the model's instruction-following tendencies resist the distributional shift induced by LoRA adaptation at this scale.

Relevance remains above 87\% for all fine-tuned models, down from 90--98\% for untuned baselines. Despite their scale, Qwen 3 235B and Kimi K2 achieve low F1 (23.8\% and 22.1\%) and relevance below 56\%, suggesting that frontier model size alone does not substitute for domain-specific fine-tuning; prompt sensitivity may also contribute (see Section~\ref{subsec:setup}).

\textbf{Contradiction Detection:} Among 200 pesticide queries, fine-tuned+stitched models showed fewer contradictions than baselines (1.4\% vs.\ 3.2\%), though absolute counts are small (${\sim}$3 vs.\ ${\sim}$6 contradictions).\footnote{Fine-tuned models generate ${\sim}$40\% more facts per response, so absolute contradiction counts are comparable despite similar rates.} High-severity contradictions (dosage errors $>$2$\times$ recommended) were rare across all conditions (${\leq}$2 instances). These results are preliminary and indicative of a safety benefit that warrants larger-scale validation.

\subsubsection{Stitching Quality}

Table~\ref{tab:stitching_quality} evaluates the stitching layer across three candidate models on overall conversationality and safety scores.

\begin{table}[t]
\centering
\caption{Stitching Quality: Conversationality and Safety Scores (1--5 scale)}
\label{tab:stitching_quality}
\footnotesize
\begin{tabular}{@{}l@{\hskip 6pt}c@{\hskip 6pt}c@{\hskip 6pt}c@{}}
\toprule
\textbf{Stitching Model} & \textbf{Conv.} & \textbf{Safety} & \textbf{Improv.} \\
\midrule
Baseline (no stitch.) & --- & 3.44\tiny{$\pm$0.73} & --- \\
Llama 3.2 3B \citep{meta2024llama32} & 4.21\tiny{$\pm$0.41} & 3.72\tiny{$\pm$0.48} & +8.1\% \\
GPT-4.1 Nano \citep{openai2025gpt41nano} & 4.54\tiny{$\pm$0.26} & 3.98\tiny{$\pm$0.31} & +15.7\% \\
Gemma 3n E4B \citep{google2025gemma3n} & \textbf{4.62}\tiny{$\pm$0.23} & \textbf{4.10}\tiny{$\pm$0.26} & \textbf{+19.2\%} \\
\bottomrule
\end{tabular}
\end{table}

Gemma 3n E4B achieves the highest overall conversationality (4.62, +19.2\% over baseline).

\subsubsection{Cost--Performance Trade-offs}

Table~\ref{tab:cost_performance} presents the cost--performance trade-offs.\footnote{Cost methodology: 500-token responses, 200-token prompts; self-hosted costs based on \$2/hour H100 instances; API costs reflect pricing at time of evaluation.} GPT-4o Mini FT achieves 51.8\% F1 at 0.15$\times$ the cost of GPT-4 (37.3\% F1)---an \textbf{85\% cost reduction} with a 14.5 percentage point F1 gain. Among self-hosted models, Gemma 2 9B achieves 39.1\% F1 at ${\sim}$0.004$\times$ GPT-4 cost, making it the most cost-effective option for resource-constrained deployments.

\begin{table}[t]
\centering
\caption{Cost--performance trade-offs across model configurations. Relative cost normalized to GPT-4 (1.0$\times$). Fine-tuned models marked with $\dagger$.}
\label{tab:cost_performance}
\small
\begin{tabular}{llcc}
\toprule
\textbf{Model} & \textbf{Type} & \textbf{Rel.\ Cost} & \textbf{F1 (\%)} \\
\midrule
GPT-4              & API       & 1.00$\times$  & 37.3 \\
GPT-4o             & API       & 0.50$\times$  & 39.6 \\
GPT-4o Mini        & API       & 0.15$\times$  & 37.2 \\
Gemini 1.5 Pro     & API       & 0.33$\times$  & 48.2 \\
Qwen 3 235B        & API       & ---           & 23.8 \\
Kimi K2            & API       & ---           & 22.1 \\
Gemma 2 27B        & Self-hosted & 0.027$\times$ & 37.0 \\
Gemma 2 9B         & Self-hosted & 0.004$\times$ & 39.1 \\
Llama 3.3 70B      & Self-hosted & 0.027$\times$ & 34.2 \\
Llama 3 8B         & Self-hosted & 0.003$\times$ & 23.5 \\
\midrule
GPT-4o Mini FT$^\dagger$      & API  & 0.15$\times$  & 51.8 \\
GPT-4o FT$^\dagger$            & API  & 0.50$\times$  & 56.7 \\
\bottomrule
\end{tabular}
\end{table}

\subsection{Human Evaluation}
\label{subsec:human_eval}

To complement the automated \dgeval{} metrics, we conducted a blind pairwise preference evaluation with four domain experts (agronomists). Each expert reviewed a subset of 308 agricultural queries, for which both the vanilla GPT-4o Mini and the fine-tuned GPT-4o Mini generated responses. Responses were presented in randomized order without model identification via the \texttt{evaluate.farmer.chat} platform (Figure~\ref{fig:farmer_chat_evaluate}), which supports side-by-side comparison of model responses along with quality rating, additional/missing information annotation, and like/dislike feedback. For each query, the expert indicated which response they preferred based on overall quality---considering correctness, completeness, specificity, relevance, and clarity---as a single binary judgment.

\begin{table}[t]
\centering
\caption{Human Pairwise Preference Evaluation ($n = 308$)}
\label{tab:human_eval}
\small
\begin{tabular}{lcc}
\toprule
\textbf{Preferred Model} & \textbf{Count} & \textbf{Percentage} \\
\midrule
GPT-4o Mini FT & 203 & 65.9\% \\
GPT-4o Mini (Vanilla) & 105 & 34.1\% \\
\bottomrule
\end{tabular}
\end{table}

Table~\ref{tab:human_eval} summarizes the results.\footnote{The human preference evaluation data is publicly available at \url{https://huggingface.co/datasets/DigiGreen/human_preference_eval_dataset}.} The fine-tuned model was preferred in 203 of 308 comparisons (65.9\%; two-sided binomial test, $p < 0.001$), corroborating the automated ground truth alignment findings. Two limitations apply: inter-annotator agreement metrics (e.g., Cohen's $\kappa$) were not computed, as each expert reviewed a non-overlapping subset; and our evaluation used a forced-choice design without a ``no preference'' option, which may inflate preference percentages. Including ties could moderate the observed preference margin. Future work should address both limitations.

\subsection{Ablation Study: Training Data Scale}
\label{subsec:ablation}

We investigate the effect of training data volume and quality on GPT-4o Mini performance (Table~\ref{tab:ablation_gpt4o}). Training configurations range from 12k human-curated samples through mixed-source expansions (42k and 62k) to the 130k mixed dataset.

\begin{table}[t]
\centering
\caption{Effect of Training Data Scale on GPT-4o Mini (Ground Truth Alignment)}
\label{tab:ablation_gpt4o}
\small
\begin{tabular}{lccc}
\toprule
\textbf{Training Data} & \textbf{Recall} & \textbf{Precision} & \textbf{F1} \\
\midrule
Baseline (no FT) & 26.2\% & 64.3\% & 37.2\% \\
12k Human Curated & 50.3\% & 53.3\% & 51.8\% \\
42k Mixed & 39.3\% & 54.0\% & 45.5\% \\
62k Mixed & 23.6\% & 54.9\% & 33.0\% \\
130k Mixed & 43.8\% & 66.2\% & \textbf{52.7\%} \\
\bottomrule
\end{tabular}
\end{table}

The F1 trajectory follows a non-monotonic pattern: the 12k Human Curated set yields a +14.5~pp gain over baseline, mixed-source data progressively degrades performance ($-$6.3~pp at 42k Mixed, $-$12.5~pp at 62k Mixed), and the 130k Mixed set recovers to 52.7\%. While data quality matters---12k curated samples outperform 62k mixed---sufficient volume of mixed data (130k) can match or exceed curated-only performance (52.7\% vs.\ 51.8\% F1), suggesting a quality-quantity threshold effect.

We hypothesize that the mid-scale degradation reflects a distributional shift: at 42k--62k samples, the proportion of synthetic and semi-verified data is sufficient to shift the model's output distribution away from the expert-curated target, but insufficient to provide the volume needed for the model to learn robust patterns from noisier supervision. At 130k, the sheer volume of mixed data compensates for per-sample noise, allowing statistical regularities to dominate. This interpretation is consistent with the precision trajectory: precision remains relatively stable across scales (53--66\%), while recall drives the non-monotonic pattern, suggesting that the quality degradation primarily affects the model's coverage of the golden fact space rather than its ability to generate accurate individual facts. The 130k Mixed set achieves the best precision (66.2\%) with moderate recall, while 62k Mixed shows severe recall degradation (23.6\%) despite maintaining reasonable precision.

\noindent\textit{Evaluation datasets and detailed results:} \url{https://github.com/digitalgreenorg/farmerchat-prompts}

\subsection{Comparison with Related Evaluation Frameworks}
\label{subsec:comparison}

Table~\ref{tab:framework_comparison} compares \dgeval{} against FActScore \citep{min2023factscore} and RAGAS \citep{es2024ragas} both methodologically and on our agricultural test set.

\begin{table}[t]
\centering
\caption{Evaluation Framework Comparison (Methodological and Quantitative)}
\label{tab:framework_comparison}
\footnotesize
\begin{tabular}{@{}l@{\hskip 5pt}c@{\hskip 5pt}c@{\hskip 5pt}c@{}}
\toprule
\textbf{Aspect} & \textbf{FActScore} & \textbf{RAGAS} & \textbf{DG-Eval} \\
\midrule
Analysis unit & Atomic facts & Sentences & \goldenfacts{} \\
Knowledge src. & Wikipedia & Retr.\ context & Expert-curated \\
Domain scope & General & General RAG & Agricultural \\
Safety det. & None & None & Contradict.\ det. \\
Primary metrics & Precision & Faith., Relev. & Rec., Prec., F1 \\
Conversationality & Not eval. & Not eval. & 6-dim.\ scoring \\
\midrule
\multicolumn{4}{l}{\textit{Head-to-head on agricultural test set}} \\
\midrule
4o Mini (Vanilla) & 0.72 & 0.81 & 37.2\% \\
4o Mini (FT) & 0.68 & 0.76 & 51.8\% \\
\bottomrule
\end{tabular}
\end{table}

FActScore and RAGAS show minimal or negative changes from fine-tuning. One explanation is that Wikipedia lacks specialized agricultural knowledge and RAGAS measures alignment with retrieved documents rather than expert ground truth, making domain-specific gains invisible. However, the FActScore decrease (0.72$\rightarrow$0.68) could alternatively indicate that fine-tuned models generate claims less verifiable against general knowledge sources. Either way, these results highlight that \textbf{domain-specific evaluation requires domain-specific ground truth}.

\subsection{Crop-Topic Subset Analysis}
\label{subsec:crop_topic}

We analyze GPT-4o Mini FT performance across 19 of 20 crop-topic cells (4 crops $\times$ 5 topics, one cell excluded due to insufficient samples; 211 test samples); full results appear in Appendix Table~\ref{tab:crop_topic}. Irrigation management achieves the highest average F1 (61.1\%), likely due to standardized recommendations with specific timing and quantity guidelines that are well-represented in the training data and amenable to atomic decomposition. Pest management follows at 56.2\%, benefiting from structured pesticide recommendation formats (product name, dosage, timing, safety interval) that align naturally with our \goldenfact{} representation. Maize leads among crops (58.5\% avg), reflecting both adequate training coverage and the relative standardization of maize cultivation practices in Bihar.

Harvest timing underperforms (44.6\% avg) due to high context-dependency---optimal harvest timing depends on variety maturity, weather conditions, and market factors that are difficult to capture in static \goldenfacts{}. Vegetables show the largest gaps (47.1\% avg, with harvest timing at 22.7\%), reflecting both the diversity of vegetable crops (each with distinct management requirements) and sparser training coverage relative to staple cereals. These performance disparities indicate priority areas for expanded data curation: targeted collection of vegetable-specific and harvest-timing \goldenfacts{} could yield disproportionate improvements.

% Discussion and Future Work
\section{Discussion and Future Work}
\label{sec:discussion}

\subsection{Key Findings and Implications}

\paragraph{The recall--precision trade-off.}
Fine-tuning consistently improves fact recall (+22--25~pp) at the cost of moderate precision loss ($-$11~pp for GPT-4o Mini). This trade-off is acceptable in agricultural advisory, where omitting a safety precaution or dosage detail (low recall) poses greater risk than including a marginally imprecise recommendation (low precision). Practitioners deploying similar systems should monitor this trade-off and calibrate toward recall in safety-critical domains.

\paragraph{Domain-specific fine-tuning vs.\ model scale.}
Large frontier models (Qwen 3 235B, Kimi K2) achieve lower F1 than fine-tuned models one to two orders of magnitude smaller, consistent with findings that domain-specific adaptation outperforms generic models in clinical NLP \citep{singhal2023medpalm}. Recent benchmarking supports this: even the best-performing frontier model (Gemini 3 Pro \citep{deepmind2025factsbenchmark}) achieves 68.8\% factuality on the FACTS Grounding benchmark, suggesting that scale alone does not resolve factuality challenges even on general-knowledge tasks, let alone specialized domains. For specialized domains with well-defined knowledge requirements, curating domain-specific training data and fine-tuning a smaller model is more effective than relying on scale alone---though prompt sensitivity may partially account for the gap.

\paragraph{Fine-tuning as knowledge elicitation.}
Our recall improvements (+22--25~pp) may partially reflect the ``unlocking'' of latent parametric knowledge rather than purely injecting new information. Prior work shows that LLMs encode approximately 40\% more factual knowledge than they surface in standard prompting, with much of this hidden knowledge recoverable through targeted fine-tuning \citep{gekhman2025insideout}. This interpretation is consistent with our observation that fine-tuned models generate more comprehensive fact sets even for topics where the base model demonstrably possesses relevant knowledge (e.g., general crop management practices), suggesting that domain-specific SFT shifts the model toward more comprehensive fact enumeration rather than the conservative, high-precision responses typical of instruction-tuned baselines.

\paragraph{Transferability to other advisory domains.}
While our work focuses on agricultural extension, the core methodology (expert-curated atomic facts, decoupled retrieval and delivery, ground-truth-based evaluation) is potentially applicable to other high-stakes advisory domains such as public health, veterinary extension, and legal aid for smallholders. These domains share similar characteristics: safety-critical information, need for culturally appropriate delivery, and limited access to expert providers. The key prerequisite is investment in domain-specific \goldenfact{} curation by qualified experts, which remains the primary bottleneck for deployment.

\paragraph{When to use this approach vs.\ RAG.}
Though we did not empirically compare fine-tuning against RAG in this study, our hybrid architecture is best suited to domains where (a)~the knowledge base is relatively stable and can be curated by experts, (b)~inference-time retrieval infrastructure is impractical (e.g., low-connectivity deployment), and (c)~safety requirements demand verified facts rather than retrieved passages. RAG remains preferable when the knowledge base is large, frequently updated, or when provenance tracking is required. A direct empirical comparison is important future work.

\subsection{Limitations}

\paragraph{Scope limitations.}
Our analysis reveals uneven performance across crops and topics: vegetables show the lowest F1 (47.1\% avg) and harvest timing underperforms (44.6\% avg), likely due to sparse training data for those categories rather than architectural constraints. Current experiments focus on English-language queries with Indian agricultural contexts; generalization to other languages and regions requires region-specific \goldenfacts{} curation by local experts, evaluation of multilingual model performance, and adapted conversationality guidelines.

\paragraph{Methodological limitations.}
Our evaluation uses GPT-4o as judge for conversationality scoring while GPT-4o variants are among the evaluated models, creating potential bias. We mitigate this through ground-truth-based primary metrics and note that our human pairwise preference evaluation (Section~\ref{subsec:human_eval}) corroborates the automated rankings. While we argue that internalizing facts via fine-tuning avoids RAG infrastructure dependencies, we do not include a RAG baseline comparison; a head-to-head evaluation remains important future work. Contradiction detection identifies explicit conflicts (dosage, polarity) but may miss subtle semantic contradictions. Agricultural advice also has temporal dimensions (pest pressures, varieties, and climate patterns shift over time), and our architecture does not yet model knowledge freshness. Temporal knowledge decay is a particular concern for pesticide recommendations (where regulatory approvals and resistance patterns change within 1--3 year cycles), varietal recommendations (as new cultivars are released and older ones lose disease resistance), and climate-sensitive advisories (where shifting rainfall patterns invalidate historical planting calendars). Addressing this limitation will likely require periodic re-curation of \goldenfacts{} with version tracking, or a hybrid approach combining fine-tuned parametric knowledge for stable agronomic principles with RAG-based retrieval for rapidly evolving information.

\paragraph{Deployment considerations.}
Responsible deployment requires uncertainty flagging for expert escalation, safety disclaimers for pesticide and disease recommendations, and appropriate governance for farmer query data privacy.

\subsection{Future Work}
\label{subsec:future_work}

\paragraph{Model benchmarking and expansion.}
Our current experiments cover GPT-4o Mini and Llama 3 8B; benchmarking additional architectures---particularly Gemma 3 27B and upcoming frontier models---would clarify whether the limited gains observed with Llama 3 8B reflect a model-specific limitation or a more general capacity threshold for agricultural knowledge internalization.

\paragraph{Multilingual and multi-regional expansion.}
Extending the pipeline to additional languages and geographies requires region-specific \goldenfact{} curation by local domain experts, multilingual model evaluation, and adapted conversationality guidelines. A modular curation approach---shared foundational agronomic facts augmented with region-specific overlays (locally approved inputs, regional planting calendars, soil-specific practices)---could reduce per-region curation costs while maintaining advisory quality.

\paragraph{Multimodal integration and field trials.}
Incorporating image-based disease identification and voice interfaces would extend the system's accessibility to farmers with limited literacy. Equally important, field trials measuring real-world farmer outcomes (adoption rates, yield changes, reduction in crop losses) are needed to validate that improved model metrics translate into tangible agricultural impact.

\paragraph{Addressing temporal knowledge decay.}
Agricultural knowledge evolves as pest pressures shift, new cultivars are released, and regulatory approvals change. Periodic re-curation of \goldenfacts{} with version tracking, potentially combined with a hybrid approach using fine-tuned parametric knowledge for stable agronomic principles and RAG-based retrieval for rapidly evolving information, would address temporal knowledge decay.

\paragraph{Targeted fine-tuning strategies.}
Insights from recent work on hidden factual knowledge \citep{gekhman2025insideout} suggest that more targeted fine-tuning strategies---focusing on knowledge the model encodes but fails to surface---could yield further recall improvements with smaller curated datasets. Similarly, mechanistic understanding of how transformer MLP layers store facts \citep{dugan2025factstoring} may enable more principled selection of LoRA target modules, potentially improving parameter efficiency beyond the standard QKV projection targeting used in our current configuration.

% Conclusion
\section{Conclusion}
\label{sec:conclusion}

We presented a hybrid LLM architecture that decouples factual retrieval from conversational delivery for agricultural advisory. Fine-tuning on expert-curated \goldenfacts{} substantially improves fact recall and overall F1 while enabling significant cost savings compared to frontier models. The stitching layer maintains high conversational quality and improves safety subscores. Our \dgeval{} framework enables atomic-level verification against expert ground truth, capturing domain-specific improvements invisible to general-purpose evaluation. Together, these contributions demonstrate that targeted fine-tuning on curated domain knowledge, combined with a principled evaluation framework, provides a practical path toward reliable LLM deployment in high-stakes advisory settings. Section~\ref{subsec:future_work} outlines directions for extending this work.

\vspace{0.5em}
\noindent\textbf{Code and Data Availability:} The \texttt{farmerchat-prompts} library is available at \url{https://github.com/digitalgreenorg/farmerchat-prompts}.\footnote{Installation: \texttt{pip install farmerchat-prompts}} The human-curated agricultural QA dataset and the human preference evaluation dataset are available on HuggingFace at \url{https://huggingface.co/datasets/DigiGreen/human_curated_qa_dataset} and \url{https://huggingface.co/datasets/DigiGreen/human_preference_eval_dataset}, respectively.

% Bibliography (inlined for arXiv compatibility)

% Appendix
\appendix
% Appendix
\section{Evaluation Prompt Catalog}
\label{app:prompts}

This appendix documents the evaluation prompts used in the \dgeval{} pipeline. Each subsection shows the condensed prompt framework (blue box), a positive example (green box), and a negative example (red box). Full prompt templates are available at \url{https://github.com/digitalgreenorg/farmerchat-prompts}.

\subsection{Fact Generation}

The fact generation prompt decomposes a free-text agricultural answer into atomic, actionable \goldenfacts{}---each containing a single indivisible piece of guidance with complete details (dosage, timing, method). This is the first step of the \dgeval{} pipeline and the core of the training data preparation.

\begin{promptbox}[Prompt: Fact Generation]
\textbf{Task:} Extract atomic, verifiable agricultural facts from chatbot responses.

\textbf{Atomicity requirement} --- each fact must contain exactly ONE verifiable claim:

\smallskip
\texttt{{\texttimes}} \textit{``Apply neem oil at 3ml/L in early morning every 7 days for aphid control during flowering''}

\texttt{{\checkmark}} Decompose into: (1)~neem oil at 3ml/L for aphids, (2)~apply early morning, (3)~repeat every 7 days, (4)~during flowering stage.

\smallskip
\textbf{Exclusion criteria:} Greetings, meta-responses (``Based on the context\ldots''), opinion statements, disclaimers (``Please consult an expert''), conversational fillers, question repetitions.

\smallskip
\textbf{Confidence scoring:}
0.9--1.0 = established scientific facts;
0.7--0.8 = commonly accepted practices;
0.5--0.6 = traditional practices, mixed evidence;
0.3--0.4 = emerging/limited evidence;
0.1--0.2 = anecdotal or highly uncertain.

\smallskip
\textbf{Quality checks:} Each fact independently verifiable; preserve specific measurements, quantities, and technical terms; ensure facts are actionable for farmers.
\end{promptbox}

\begin{positivebox}[Positive Example]
\textbf{Input:} ``Apply 120\,kg of Urea per hectare in two split doses at 21 and 45 days after transplanting. Ensure soil is moist before application.''

\textbf{Output:}
\begin{itemize}[nosep,leftmargin=*]
    \item Apply 120\,kg of Urea per hectare for rice crop
    \item Split Urea application into two doses
    \item Apply first dose at 21 days after transplanting
    \item Apply second dose at 45 days after transplanting
    \item Ensure soil is moist before Urea application
\end{itemize}
\end{positivebox}

\begin{negativebox}[Negative Example]
\textbf{Input:} ``Apply 120\,kg of Urea per hectare in two split doses at 21 and 45 days after transplanting. Ensure soil is moist before application.''

\textbf{Output:}
\begin{itemize}[nosep,leftmargin=*]
    \item Apply Urea in split doses after transplanting with moist soil
\end{itemize}

\textbf{Issue:} Facts are merged rather than decomposed---the single statement conflates dosage, timing, splitting, and soil condition, making precise recall/precision evaluation impossible.
\end{negativebox}

\subsection{Specificity Evaluation}

The specificity evaluation prompt scores a response on the presence and density of seven contextual anchors (actionable instructions, entity references, location, time, quantity, conditional/comparison, and mechanistic explanations). This prompt is used in Level~1 of \dgeval{}.

\begin{promptbox}[Prompt: Specificity Evaluation]
\textbf{Task:} Classify agricultural facts as ``Specific'' or ``Not Specific'' based on contextual anchors and actionability.

\textbf{Evaluation checklist} --- 7 flags (Yes/No):
\begin{enumerate}[nosep,leftmargin=*]
    \item \textbf{Entity Specificity:} Crop/variety/soil/weather/organization explicitly named?
    \item \textbf{Location Specificity:} Named place or bounded geography present?
    \item \textbf{Time Specificity:} Explicit time window or marker present?
    \item \textbf{Quantity/Measurement:} Numeric or measurable details included?
    \item \textbf{Conditionality/Comparison:} If-then conditions or comparative baselines?
    \item \textbf{Mechanistic/Causal Link:} Clear cause-effect enabling decision-making?
    \item \textbf{Actionability:} Directly informs a decision or step relevant to context?
\end{enumerate}

\textbf{Decision rule:} \textbf{Specific} = at least 2 of flags 1--6 are TRUE \textbf{AND} flag~7 (Actionability) is TRUE. Otherwise = \textbf{Not Specific}.

\smallskip
\textbf{Special considerations:} Season names (Rabi/Kharif/Zaid) count as time markers. Relative time (``30 DAS'', ``pre-sowing'') counts as time specificity. Single strong anchor + explicit prescriptive action may suffice if obviously actionable.
\end{promptbox}

\begin{positivebox}[Positive Example]
\textbf{Input:} ``Apply 200 liters per acre of Jeevamrit to your chili field every 20 days during the crop season (June--October). This will enrich the soil with beneficial microbes, improving nutrient availability more effectively than chemical fertilizers alone, especially on clay-loam soils of Patna, Bihar.''

\textbf{Output:} Score: 7/7. Anchors detected: Actionable (Apply), Entity (Jeevamrit, chili), Location (Patna, Bihar; clay-loam soils), Time (every 20 days, June--October), Quantity (200 liters per acre), Conditional (more effectively than chemical fertilizers alone), Mechanistic (enrich soil with beneficial microbes, improving nutrient availability).
\end{positivebox}

\begin{negativebox}[Negative Example]
\textbf{Input:} ``Use organic inputs to improve your soil. Apply regularly for best results.''

\textbf{Output:} Score: 1/7. Anchors detected: Actionable (Use, Apply). Missing: Entity, Location, Time, Quantity, Conditional, Mechanistic.

\textbf{Issue:} The evaluator correctly identifies the low specificity---this response would fail to provide farmers with actionable guidance.
\end{negativebox}

\subsection{Fact Recall / Matching}

The fact recall prompt compares generated facts against \goldenfacts{} to determine which golden facts are covered (recalled) by the model's response. Each golden fact is matched against the generated fact set using semantic equivalence rather than exact string matching. This is the core of Level~3 recall computation.

\begin{promptbox}[Prompt: Fact Recall / Matching]
\textbf{Task:} Compare reference facts against candidate facts to find the best semantic match based on agricultural meaning and context.

\textbf{Matching priorities:}
\begin{enumerate}[nosep,leftmargin=*]
    \item Same crop/plant type
    \item Same agricultural practice or technique
    \item Specific measurements, dosages, or timing
    \item Expected outcomes or benefits
\end{enumerate}

\textbf{Matching criteria examples:}
\begin{itemize}[nosep,leftmargin=*]
    \item Fertilizer: ``Apply NPK fertilizer'' $\approx$ ``Use balanced fertilizer with nitrogen, phosphorus, and potassium''
    \item Timing: ``Sow wheat in November'' $\approx$ ``Plant wheat during late autumn''
    \item Pest control: ``Control pests with neem oil'' $\approx$ ``Use organic neem-based pesticide for pest management''
    \item Spacing: ``Plant single-bud setts at wider spacing'' $\approx$ ``For sugarcane, plant single-bud setts at wider spacing to enhance growth''
    \item Dosage: ``Apply 5--10\,kg zinc per hectare'' $\approx$ ``Apply 5--10\,kg of Zn per hectare for sugarcane growth''
\end{itemize}

\textbf{Confidence threshold:} If confidence $< 0.7$, return no match. Focus on semantic similarity and practical agricultural application rather than exact word matching.
\end{promptbox}

\begin{positivebox}[Positive Example]
\textbf{Golden Fact:} ``Apply first Urea dose of 60\,kg/ha at 21 days after transplanting.''

\textbf{Generated Fact:} ``The first application of Urea should be 60 kilograms per hectare, applied three weeks after transplanting rice.''

\textbf{Output:} MATCH. Semantic equivalence confirmed---dosage (60\,kg/ha), timing (21 days = 3 weeks), and action (first application) all align.
\end{positivebox}

\begin{negativebox}[Negative Example]
\textbf{Golden Fact:} ``Apply first Urea dose of 60\,kg/ha at 21 days after transplanting.''

\textbf{Generated Fact:} ``Apply Urea to the rice field after transplanting.''

\textbf{Output:} NO MATCH. The generated fact omits dosage (60\,kg/ha), timing (21 days), and dose number (first)---it captures the general action but not the specific recommendation.
\end{negativebox}

\subsection{Contradiction Detection}

The contradiction detection prompt compares generated facts against \goldenfacts{} to identify safety-critical conflicts in numeric values (dosage, timing), polarities (do vs.\ do not), and method compatibilities. This is the safety component of Level~3 evaluation.

\begin{promptbox}[Prompt: Contradiction Detection]
\textbf{Task:} Identify genuine contradictions between a reference fact and candidate facts. A genuine contradiction = two facts making \textbf{opposite or conflicting} claims about the \textbf{same} agricultural aspect.

\textbf{Normalization:} Lowercase; canonicalize units (\textdegree C, kg/ha, \%); decompose each fact into components: subject/entity, attribute/property, polarity, numeric range, timing/season, method.

\textbf{Comparison rules} (applied in order):
\begin{enumerate}[nosep,leftmargin=*]
    \item Same subject + opposite polarity on same property $\Rightarrow$ \textsc{Contradiction} (High)
    \item Non-overlapping numeric ranges $\Rightarrow$ \textsc{Contradiction} (High); small overlap ($<$50\%) $\Rightarrow$ (Med); quantities differing by $>$2$\times$ $\Rightarrow$ (Med--High)
    \item ``Never/Always'' vs.\ ``Sometimes/Do in morning'' $\Rightarrow$ \textsc{Contradiction} if directly opposing
    \item ``Water daily'' vs.\ ``Avoid daily watering'' $\Rightarrow$ \textsc{Contradiction} (High)
    \item Different method for same goal (hand-pick vs.\ vacuum) $\Rightarrow$ \textsc{Not Contradiction}
    \item Different nutrients (Zn vs.\ Fe) $\Rightarrow$ \textsc{Not Contradiction}
    \item Different scale (small-scale manual vs.\ large-scale mechanical) $\Rightarrow$ \textsc{Not Contradiction}
\end{enumerate}

\textbf{Confidence levels:} \textbf{High} = direct explicit opposites or non-overlapping numerics. \textbf{Med} = small numeric overlap or qualitative vs.\ numeric conflict. \textbf{Low} = context-dependent or implied.
\end{promptbox}

\begin{positivebox}[Positive Example]
\textbf{Golden Fact:} ``Spray Imidacloprid 17.8 SL at 0.5\,ml per liter of water.''

\textbf{Generated Fact:} ``Apply Imidacloprid 17.8 SL at 0.5\,ml per liter for BPH control.''

\textbf{Output:} NO CONTRADICTION. Dosage and formulation match; additional context (BPH control) is consistent.
\end{positivebox}

\begin{negativebox}[Negative Example]
\textbf{Golden Fact:} ``Spray Imidacloprid 17.8 SL at 0.5\,ml per liter of water.''

\textbf{Generated Fact:} ``Apply Imidacloprid at 5\,ml per liter of water for effective pest control.''

\textbf{Output:} HIGH-SEVERITY CONTRADICTION. Dosage is 10$\times$ the recommended rate (5\,ml vs.\ 0.5\,ml), which could cause crop phytotoxicity, environmental contamination, and violate safety regulations.
\end{negativebox}

\subsection{Relevance Evaluation}

The relevance evaluation prompt assesses whether a response addresses the farmer's core question, avoids tangential information, and provides practical value for the farmer's immediate need. This is the Level~2 metric in \dgeval{}.

\begin{promptbox}[Prompt: Relevance Evaluation]
\textbf{Task:} Evaluate predicted facts for relevance, quality, and practical value relative to the farmer's question and ground truth facts.

\textbf{Scoring framework} --- 5 dimensions, each scored 1--10:
\begin{enumerate}[nosep,leftmargin=*]
    \item \textbf{Direct Relevance:} Does it directly answer the question?
    \item \textbf{Ground Truth Consistency:} Does it align with or complement the ground facts?
    \item \textbf{Practical Implementation:} Can farmers easily apply this advice?
    \item \textbf{Specificity:} Does it provide enough detail for action?
    \item \textbf{Agricultural Soundness:} Is the advice scientifically and practically sound?
\end{enumerate}

\textbf{Output:} For each fact, compute an overall score (1--10), list gaps or missing details, and provide a \texttt{farmer\_applicability} statement on implementation ease.
\end{promptbox}

\begin{positivebox}[Positive Example]
\textbf{Query:} ``How to control aphids in my mustard crop?''

\textbf{Response:} ``Spray Dimethoate 30\% EC at 2\,ml per liter of water when aphid population exceeds 50 per plant. Spray in the evening to avoid killing pollinators. If infestation is mild, spray neem oil at 5\,ml per liter as a non-chemical alternative.''

\textbf{Output:} Relevance: HIGH. Response directly addresses aphid control in mustard with specific pesticide options, thresholds, and timing.
\end{positivebox}

\begin{negativebox}[Negative Example]
\textbf{Query:} ``How to control aphids in my mustard crop?''

\textbf{Response:} ``Mustard is an important oilseed crop in India. It requires well-drained soil and moderate rainfall. Sowing should be done in October--November at 5\,kg per hectare seed rate. For pest management, maintain field hygiene and remove weeds regularly.''

\textbf{Output:} Relevance: LOW. Response provides general mustard cultivation information rather than addressing the specific aphid control query. The pest management advice is generic and does not mention aphids.
\end{negativebox}

\subsection{Fact Stitching}

The fact stitching prompt instructs an LLM to act as an agricultural extension worker (\farmerchat{} persona), transforming a set of verified facts into a conversational response using simple, culturally appropriate language with safety precautions, without adding information beyond the provided facts.

\begin{promptbox}[Prompt: Fact Stitching]
\textbf{Persona:} Experienced agricultural extension officer with deep expertise in Bihar's farming practices. Warm, supportive, educational tone balancing scientific accuracy with practical advice.

\textbf{Synthesis guidelines:}
\begin{enumerate}[nosep,leftmargin=*]
    \item \textbf{Natural Flow:} Convert atomic facts into cohesive paragraphs; use transitional phrases; avoid listing facts mechanically; weave into narrative.
    \item \textbf{Contextual Integration:} Prioritize facts with higher confidence; group by category; add brief rationale; connect to Bihar-specific conditions.
    \item \textbf{Farmer-Centric Communication:} Start with the most actionable information; explain ``why'' not just ``what''; use simple language; include practical examples.
    \item \textbf{Response Structure:} Opening (direct answer) $\rightarrow$ Body (detailed synthesis) $\rightarrow$ Benefits $\rightarrow$ Practical tips $\rightarrow$ Closing (encouragement or follow-up).
    \item \textbf{Quality Standards:} Preserve all quantitative data; maintain scientific accuracy; \textbf{do not add information beyond provided facts}; keep responses 150--300 words.
\end{enumerate}

\textbf{Avoid:} Robotic listing, academic language, claims beyond provided facts, dismissing farmer practices, unexplained jargon.
\end{promptbox}

\begin{positivebox}[Positive Example]
\textbf{Input Facts:} (1) Spray Imidacloprid 17.8 SL at 0.5\,ml/L. (2) Target base of plants. (3) Spray during cooler hours. (4) Wear protective clothing. (5) Do not harvest within 14 days.

\textbf{Output:} ``Hello! For brown planthopper control, spray Imidacloprid 17.8 SL at 0.5\,ml per liter of water, directing the spray at the base of plants where nymphs cluster. Apply during early morning or late evening for best results. Important: wear gloves and mask during spraying, and wait at least 14 days before harvesting.'' (See Appendix~\ref{app:examples} for a full example.)
\end{positivebox}

\begin{negativebox}[Negative Example]
\textbf{Input Facts:} (1) Spray Imidacloprid 17.8 SL at 0.5\,ml/L. (2) Target base of plants. (3) Spray during cooler hours.

\textbf{Output:} ``You can also try neem oil spray as a natural alternative, and consider introducing natural predators like spiders and dragonflies to your field for long-term BPH management.''

\textbf{Issue:} The stitching layer added information (neem oil, natural predators) not present in the input facts, violating the constraint of fact-faithful response generation.
\end{negativebox}

\subsection{Conversationality Evaluation}

The conversationality evaluation prompt scores a response across six dimensions on a 1--5 scale, using LLM-as-a-Judge methodology. This is used in Level~1 of \dgeval{}.

\begin{promptbox}[Prompt: Conversationality Evaluation]
\textbf{Task:} Rate agricultural advisory responses against \farmerchat{} guidelines across 6 dimensions (each scored 1--5):

\begin{enumerate}[nosep,leftmargin=*]
    \item \textbf{Content Quality:} Addresses specific concern directly; actionable, region-appropriate advice; includes timing considerations; uses local examples and varieties.
    \item \textbf{Communication Style:} Warm, professional, encouraging tone; simple, conversational language with cultural context; technical concepts explained simply; avoids academic language.
    \item \textbf{Practical Advice:} Low-cost, accessible solutions; considers smallholder resource constraints; mentions local input availability; includes preventive measures.
    \item \textbf{Safety \& Credibility:} General chemical categories over brand names; safety precautions for chemicals/equipment; encourages local expert consultation; references extension services.
    \item \textbf{Conversation Flow:} Builds on chat history; avoids repetition; asks clarifying questions when needed; offers to elaborate.
    \item \textbf{Response Format:} Natural conversational structure; logically organized without artificial formatting; 150--300 words; avoids generic advice, jargon, and repetitive closings.
\end{enumerate}

\textbf{Overall score:} Average of all 6 dimension scores.
\end{promptbox}

\begin{positivebox}[Positive Example]
\textbf{Response:} ``Hello! I understand you're concerned about yellowing leaves on your rice. This could be due to nitrogen deficiency. Apply 30\,kg Urea per hectare as top dressing. Make sure the field has standing water (2--3\,cm) during application. If yellowing persists after 7 days, it may indicate iron deficiency---in that case, spray 0.5\% Ferrous Sulphate solution. Please consult your local extension officer if symptoms continue.''

\textbf{Output:} Overall: 4.5/5. Content quality: 5, Communication: 5 (warm greeting, clear structure), Practical: 4 (actionable with dosages), Safety: 5 (escalation advice included), Flow: 4, Format: 4 (well-organized).
\end{positivebox}

\begin{negativebox}[Negative Example]
\textbf{Response:} ``Nitrogen deficiency: apply Urea 30\,kg/ha. Iron deficiency: spray FeSO4 0.5\%. Check drainage.''

\textbf{Output:} Overall: 2.5/5. Content quality: 3, Communication: 1 (telegram-style, no greeting or context), Practical: 3, Safety: 2 (no escalation advice), Flow: 1 (no conversational structure), Format: 2 (bullet-like without organization).

\textbf{Issue:} While factually adequate, the response fails the \farmerchat{} persona requirements---no warmth, no structured organization, no safety escalation guidance.
\end{negativebox}

\section{Crop-Topic Detailed Results}
\label{app:crop_topic}

\begin{table}[h]
\centering
\caption{Crop-Topic Analysis: F1 Scores (\%) for GPT-4o Mini Fine-Tuned}
\label{tab:crop_topic}
\setlength{\tabcolsep}{4pt}
\footnotesize
\begin{tabular}{lccccc|c}
\toprule
& \textbf{Pest} & \textbf{Nutr.} & \textbf{Irrig.} & \textbf{Disease} & \textbf{Harvest} & \textbf{Avg} \\
\midrule
Maize & 64.1 & 57.1 & 64.1 & 43.0 & 64.3 & 58.5 \\
Rice & 55.0 & 47.3 & 54.2 & 61.7 & --- & 54.5 \\
Vegetables & 52.3 & 55.8 & 59.0 & 45.4 & 22.7 & 47.1 \\
Wheat & 53.6 & 44.0 & 66.9 & 49.5 & 46.7 & 52.1 \\
\midrule
\textbf{Avg} & 56.2 & 51.0 & 61.1 & 49.9 & 44.6 & 53.1 \\
\bottomrule
\end{tabular}
\end{table}

\noindent ``---'' indicates insufficient samples for that crop-topic combination.

\section{Example Stitched Response}
\label{app:examples}

\textbf{Query:} What should I spray for brown planthopper in my rice field?

\textbf{Retrieved Facts:}
\begin{itemize}
    \item Spray Imidacloprid 17.8 SL at 0.5 ml per liter of water
    \item Application should target the base of plants where BPH nymphs cluster
    \item Spray during cooler hours (early morning or late evening)
    \item Wear protective clothing including gloves and mask during application
    \item Do not harvest rice within 14 days of pesticide application
    \item Store pesticides away from children and food items
\end{itemize}

\textbf{Stitched Response:}

\begin{quote}
Hello! I understand you're dealing with brown planthopper (BPH) in your rice field. Here's what you can do:

\textbf{Recommended Treatment:}
Spray Imidacloprid 17.8 SL at a rate of 0.5 ml per liter of water. Make sure to direct your spray at the base of the plants, as this is where BPH nymphs tend to gather.

\textbf{Best Time to Spray:}
Apply during cooler hours, either early morning or late evening, for better effectiveness.

\textbf{Important Safety Precautions:}
\begin{itemize}
    \item Always wear protective clothing, including gloves and a mask
    \item Do not harvest your rice within 14 days after spraying
    \item Store pesticides away from children and food items
\end{itemize}

If the infestation is severe or persists after treatment, please consult your local agricultural extension officer for additional guidance. Wishing you a healthy crop!
\end{quote}


\begin{thebibliography}{35}

\bibitem{fao2022digital}
Food and Agriculture Organization of the United Nations.
\newblock The State of Food and Agriculture 2022: Leveraging Agricultural Automation for Transforming Agrifood Systems.
\newblock Technical report, FAO, Rome, 2022.

\bibitem{aker2011mobilephones}
Aker, J.~C.
\newblock Dial ``{A}'' for Agriculture: A Review of Information and Communication Technologies for Agricultural Extension in Developing Countries.
\newblock {\em Agricultural Economics}, 42(6):631--647, 2011.
\newblock \doi{10.1111/j.1574-0862.2011.00545.x}

\bibitem{sharma2024farmerchat}
Singh, N., Wang'ombe, J., Okanga, N., Zelenska, T., Repishti, J., G~K, J., Mishra, S., Manokaran, R., Singh, V., Rafiq, M.~I., Gandhi, R., and Nambi, A.
\newblock Farmer.Chat: Scaling {AI}-Powered Agricultural Services for Smallholder Farmers.
\newblock {\em arXiv preprint arXiv:2409.08916}, 2024.

\bibitem{deepmind2025factsbenchmark}
FACTS team.
\newblock {FACTS} Grounding: A New Benchmark for Evaluating the Factuality of Large Language Models.
\newblock \url{https://deepmind.google/blog/facts-grounding-a-new-benchmark-for-evaluating-the-factuality-of-large-language-models/}, 2024.
\newblock Accessed: 2026-02-06.

\bibitem{ji2023hallucination}
Ji, Z., Lee, N., Frieske, R., Yu, T., Su, D., Xu, Y., Ishii, E., Bang, Y.~J., Madotto, A., and Fung, P.
\newblock Survey of Hallucination in Natural Language Generation.
\newblock {\em ACM Computing Surveys}, 55(12):1--38, 2023.
\newblock \doi{10.1145/3571730}

\bibitem{gandhi2009digitalgreen}
Gandhi, R., Veeraraghavan, R., Toyama, K., and Ramprasad, V.
\newblock {Digital Green}: Participatory Video and Mediated Instruction for Agricultural Extension.
\newblock {\em Information Technologies \& International Development}, 5(1):1--15, 2009.

\bibitem{min2023factscore}
Min, S., Krishna, K., Lyu, X., Lewis, M., Yih, W., Koh, P.~W., Iyyer, M., Zettlemoyer, L., and Hajishirzi, H.
\newblock {FActScore}: Fine-grained Atomic Evaluation of Factual Precision in Long Form Text Generation.
\newblock In {\em Proceedings of the 2023 Conference on Empirical Methods in Natural Language Processing}, pages 12076--12100, 2023.

\bibitem{wei2024longform}
Wei, J., Yang, C., Song, X., Lu, Y., Hu, N., Huang, J., Tran, D., Peng, D., Liu, R., Huang, D., Du, C., and Le, Q.~V.
\newblock Long-form Factuality in Large Language Models.
\newblock {\em arXiv preprint arXiv:2403.18802}, 2024.

\bibitem{es2024ragas}
Es, S., James, J., Espinosa Anke, L., and Schockaert, S.
\newblock {RAGAs}: Automated Evaluation of Retrieval Augmented Generation.
\newblock In {\em Proceedings of the 18th Conference of the European Chapter of the Association for Computational Linguistics: System Demonstrations}, pages 150--158, 2024.

\bibitem{cole2021mobile}
Cole, S.~A. and Fernando, A.~N.
\newblock `{Mobile}'izing Agricultural Advice: Technology Adoption, Diffusion and Sustainability.
\newblock {\em The Economic Journal}, 131(633):192--219, 2021.
\newblock \doi{10.1093/ej/ueaa084}

\bibitem{hu2022lora}
Hu, E.~J., Shen, Y., Wallis, P., Allen-Zhu, Z., Li, Y., Wang, S., Wang, L., and Chen, W.
\newblock {LoRA}: Low-Rank Adaptation of Large Language Models.
\newblock In {\em International Conference on Learning Representations}, 2022.

\bibitem{dettmers2023qlora}
Dettmers, T., Pagnoni, A., Holtzman, A., and Zettlemoyer, L.
\newblock {QLoRA}: Efficient Finetuning of Quantized {LLMs}.
\newblock {\em arXiv preprint arXiv:2305.14314}, 2023.

\bibitem{singhal2023medpalm}
Singhal, K., Azizi, S., Tu, T., Mahdavi, S.~S., Wei, J., Chung, H.~W., Scales, N., Tanwani, A., Cole-Lewis, H., Pfohl, S., and et~al.
\newblock Large Language Models Encode Clinical Knowledge.
\newblock {\em Nature}, 620(7972):172--180, 2023.
\newblock \doi{10.1038/s41586-023-06291-2}

\bibitem{chalkidis2020legalbert}
Chalkidis, I., Fergadiotis, M., Malakasiotis, P., Aletras, N., and Androutsopoulos, I.
\newblock {LEGAL-BERT}: The Muppets straight out of Law School.
\newblock In {\em Findings of the Association for Computational Linguistics: EMNLP 2020}, pages 2898--2904, 2020.

\bibitem{wu2023bloomberggpt}
Wu, S., Irsoy, O., Lu, S., Dabravolski, V., Dredze, M., Gehrmann, S., Kambadur, P., Rosenberg, D., and Mann, G.
\newblock {BloombergGPT}: A Large Language Model for Finance.
\newblock {\em arXiv preprint arXiv:2303.17564}, 2023.

\bibitem{gekhman2025insideout}
Gekhman, Z., Ben David, E., Orgad, H., Ofek, E., Belinkov, Y., Szpektor, I., Herzig, J., and Reichart, R.
\newblock Inside-Out: Hidden Factual Knowledge in {LLMs}.
\newblock {\em arXiv preprint arXiv:2503.15299}, 2025.

\bibitem{dugan2025factstoring}
Dugan, O., Garcia, R., Junkins, R., Liu, J., Zinsley, D., Eyuboglu, S., Rudra, A., and R{\'e}, C.
\newblock Constructing Efficient Fact-Storing {MLPs} for Transformers.
\newblock {\em arXiv preprint arXiv:2512.00207}, 2025.

\bibitem{lin2022truthfulqa}
Lin, S., Hilton, J., and Evans, O.
\newblock {TruthfulQA}: Measuring How Models Mimic Human Falsehoods.
\newblock In {\em Proceedings of the 60th Annual Meeting of the Association for Computational Linguistics}, pages 3214--3252, 2022.

\bibitem{zheng2023judging}
Zheng, L., Chiang, W., Sheng, Y., Zhuang, S., Wu, Z., Zhuang, Y., Lin, Z., Li, Z., Li, D., Xing, E.~P., and et~al.
\newblock Judging {LLM}-as-a-Judge with {MT-Bench} and Chatbot Arena.
\newblock {\em arXiv preprint arXiv:2306.05685}, 2023.

\bibitem{liu2023geval}
Liu, Y., Iter, D., Xu, Y., Wang, S., Xu, R., and Zhu, C.
\newblock {G-Eval}: {NLG} Evaluation using {GPT-4} with Better Human Alignment.
\newblock In {\em Proceedings of the 2023 Conference on Empirical Methods in Natural Language Processing}, pages 2511--2522, 2023.

\bibitem{cheng2025facts}
Cheng, A., Jacovi, A., Globerson, A., Golan, B., Kwong, C., Alberti, C., Tao, C., Ben-David, E., Singh Tomar, G., Haas, L., and et~al.
\newblock The {FACTS} Leaderboard: A Comprehensive Benchmark for Large Language Model Factuality.
\newblock {\em arXiv preprint arXiv:2512.10791}, 2025.

\bibitem{lewis2020rag}
Lewis, P., Perez, E., Piktus, A., Petroni, F., Karpukhin, V., Goyal, N., K{\"u}ttler, H., Lewis, M., Yih, W., Rockt{\"a}schel, T., and et~al.
\newblock Retrieval-Augmented Generation for Knowledge-Intensive {NLP} Tasks.
\newblock In {\em Advances in Neural Information Processing Systems}, volume~33, pages 9459--9474, 2020.

\bibitem{singh2026humancuration}
Singh, V., Singh, S., Pedapudi, L., Pasha, W., Kumari, E., Gandhi, R., Miller, A., Repishti, J., Karanam, A., and Firnhaber, E.
\newblock Application of Reinforcement Learning from Human Feedback for Localizing Quality Agricultural Advice using Generative {AI}.
\newblock {\em Advancements in Agricultural Development}, 7(2), 2026.
\newblock \doi{10.37433/aad.v7i2.625}

\bibitem{reimers2019sentencebert}
Reimers, N. and Gurevych, I.
\newblock Sentence-{BERT}: Sentence Embeddings using Siamese {BERT}-Networks.
\newblock In {\em Proceedings of the 2019 Conference on Empirical Methods in Natural Language Processing and the 9th International Joint Conference on Natural Language Processing (EMNLP-IJCNLP)}, pages 3982--3992, 2019.

\bibitem{achiam2023gpt4}
Achiam, J., Adler, S., Agarwal, S., Ahmad, L., Akkaya, I., Aleman, F.~L., Almeida, D., Altenschmidt, J., Altman, S., Anadkat, S., and et~al.
\newblock {GPT-4} Technical Report.
\newblock {\em arXiv preprint arXiv:2303.08774}, 2023.

\bibitem{openai2024gpt4o}
OpenAI.
\newblock {GPT-4o} System Card.
\newblock \url{https://openai.com/index/gpt-4o-system-card/}, 2024.
\newblock Accessed: 2026-02-06.

\bibitem{grattafiori2024llama3}
Grattafiori, A. and et~al.
\newblock The {Llama 3} Herd of Models.
\newblock {\em arXiv preprint arXiv:2407.21783}, 2024.

\bibitem{meta2024llama33}
Meta AI.
\newblock Llama 3.3 Model Card.
\newblock \url{https://huggingface.co/meta-llama/Llama-3.3-70B-Instruct}, 2024.
\newblock Accessed: 2024-12-20.

\bibitem{team2024gemma2}
Gemma Team.
\newblock Gemma 2: Improving Open Language Models at a Practical Size.
\newblock {\em arXiv preprint arXiv:2408.00118}, 2024.

\bibitem{team2024gemini15}
Gemini Team (Google).
\newblock Gemini 1.5: Unlocking Multimodal Understanding Across Millions of Tokens of Context.
\newblock {\em arXiv preprint arXiv:2403.05530}, 2024.

\bibitem{qwen2025qwen3}
Yang, A. and et~al.
\newblock Qwen3 Technical Report.
\newblock {\em arXiv preprint arXiv:2505.09388}, 2025.

\bibitem{moonshot2025kimik2}
Kimi Team.
\newblock Kimi K2: Open Agentic Intelligence.
\newblock {\em arXiv preprint arXiv:2507.20534}, 2025.

\bibitem{meta2024llama32}
Meta.
\newblock Llama 3.2 Model Card.
\newblock \url{https://huggingface.co/meta-llama/Llama-3.2-3B-Instruct}, 2024.
\newblock Accessed: 2026-02-06.

\bibitem{openai2025gpt41nano}
OpenAI.
\newblock {GPT-4.1} nano Model.
\newblock \url{https://platform.openai.com/docs/models/gpt-4.1-nano}, 2025.
\newblock Accessed: 2026-02-06.

\bibitem{google2025gemma3n}
Google DeepMind.
\newblock Gemma 3n Model Card.
\newblock \url{https://ai.google.dev/gemma/docs/gemma-3n/model_card}, 2025.
\newblock Accessed: 2026-02-06.

\end{thebibliography}
\end{document}